\documentclass[conference]{IEEEtran}
\IEEEoverridecommandlockouts

\usepackage{cite}
\usepackage{amsmath,amssymb,amsfonts}
\usepackage{graphicx}
\usepackage{textcomp}
\usepackage{xcolor}
\usepackage{subfigure}
\usepackage{colortbl}
\usepackage{algorithm}
\usepackage{algpseudocode}
\usepackage{booktabs}
\usepackage{multirow}
\usepackage{makecell}
\usepackage{pifont}
\usepackage{hyperref}
\usepackage{marvosym}

\definecolor{mygray}{gray}{.9}
\definecolor{myblue}{RGB}{236,250,255}

\newtheorem{theorem}{Theorem}
\newtheorem{assumption}{Assumption}
\def\BibTeX{{\rm B\kern-.05em{\sc i\kern-.025em b}\kern-.08em
    T\kern-.1667em\lower.7ex\hbox{E}\kern-.125emX}}

\begin{document}

\title{FedSKC: Federated Learning with Non-IID Data via Structural Knowledge Collaboration}

\author{
\IEEEauthorblockN{Huan Wang\textsuperscript{1}, Haoran Li\textsuperscript{1,4}, Huaming Chen\textsuperscript{2}, Jun Yan\textsuperscript{1}, Lijuan Wang\textsuperscript{3}, Jiahua Shi\textsuperscript{4}, Shiping Chen\textsuperscript{5}, Jun Shen\textsuperscript{1, \Letter}}
\IEEEauthorblockA{
\textsuperscript{1}\textit{School of Computing and Information Technology, University of Wollongong, Wollongong, Australia} \\
\textsuperscript{2}\textit{School of Electrical and Computer Engineering, The University of Sydney, Sydney, Australia} \\
\textsuperscript{3}\textit{School of Cyber Engineering, Xidian University, Xi'an, China} \\
\textsuperscript{4}\textit{QLD Alliance for Agriculture and Food, The University of Queensland, Brisbane, Australia} \\
\textsuperscript{5}\textit{Principal Research Scientist, CSIRO Data61, Eveleigh, Australia} \\
hw226@uowmail.edu.au, hl644@uowmail.edu.au, huaming.chen@sydney.edu.au, jyan@uow.edu.au, \\
ljwang@xidian.edu.cn, jiahua.shi@uq.edu.au, shiping.chen@data61.csiro.au, jshen@uow.edu.au}
}

\maketitle

\begin{abstract}
With the advancement of edge computing, federated learning (FL) displays a bright promise as a privacy-preserving collaborative learning paradigm. However, one major challenge for FL is the data heterogeneity issue, which refers to the biased labeling preferences among multiple clients, negatively impacting convergence and model performance. Most previous FL methods attempt to tackle the data heterogeneity issue locally or globally, neglecting underlying class-wise structure information contained in each client. 
In this paper, we first study how data heterogeneity affects the divergence of the model and decompose it into local, global, and sampling drift sub-problems. To explore the potential of using intra-client class-wise structural knowledge in handling these drifts, we thus propose Federated Learning with Structural Knowledge Collaboration (FedSKC). The key idea of FedSKC is to extract and transfer domain preferences from inter-client data distributions, offering diverse class-relevant knowledge and a fair convergent signal. 
FedSKC comprises three components: \romannumeral1) local contrastive learning, to prevent weight divergence resulting from local training; \romannumeral2) global discrepancy aggregation, which addresses the parameter deviation between the server and clients; \romannumeral3) global period review, correcting for the sampling drift introduced by the server randomly selecting devices. 
We have theoretically analyzed FedSKC under non-convex objectives and empirically validated its superiority through extensive experimental results. Our code is at \href{https://github.com/hwang52/FedSKC}{https://github.com/hwang52/FedSKC}.
\end{abstract}


\section{Introduction}
Federated Learning (FL) involves the collaborative training of a global model by periodically aggregating model parameters from multiple privacy-preserving decentralized clients~\cite{McMahan17}. It enables clients to collectively train a global model without actual data sharing, facilitating real-world applications across multiple domains (\textit{e.g.}, autonomous driving~\cite{9827020}, recommendation system~\cite{yu2023untargeted}). 
In a typical FL scenario, each client trains a local model using their private data, sends the model updates to the server, then server performs a weight aggregation (depends on the amount of data) to get a new global model~\cite{McMahan17,kairouz2021advances}.

Despite the promising advancements made by FL in private data modeling, the non-independent and identically distributed (\textit{non-iid}) data is still an outstanding challenge drawing lots of attention~\cite{9835537,kairouz2021advances,li2020federated,tan2023federated,ZHU2021371}. 
Data from different clients often originates from various source data, leading to \emph{local drift} among clients. Moreover, the data distribution for each client is highly inconsistent, causing local optimization to deviate from the global optimum and resulting in \emph{global drift}. Furthermore, optimization of communication is compromised by the randomness of partial clients participation in each round, introduces \emph{sampling drift} and has a particularly destabilizing effect in FL.

\begin{figure}[tbp]
    \centering
    \includegraphics[width=1.0\linewidth]{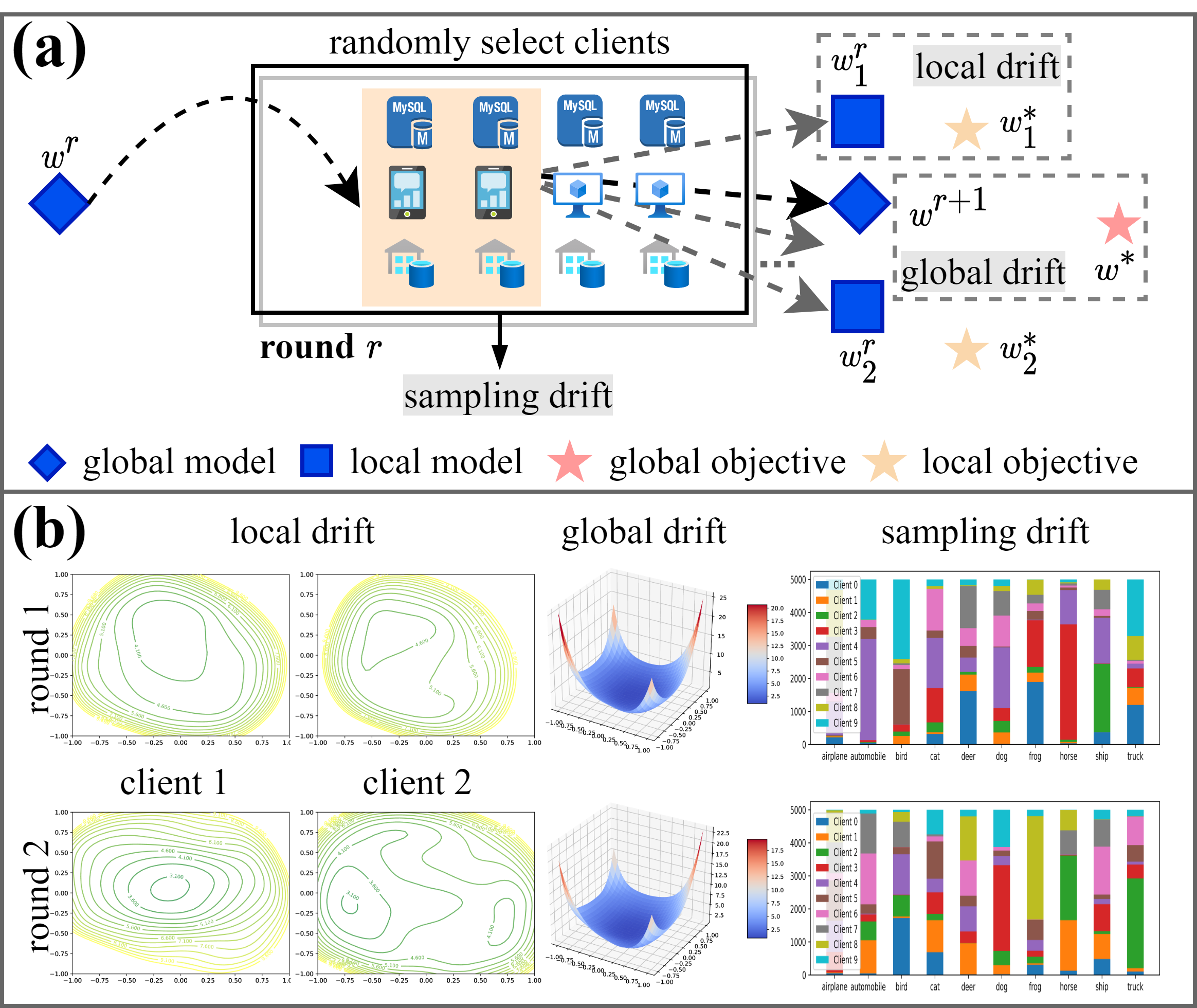}
    \caption{\textbf{(a)} An example of various drift sub-problems (local, global, and sampling drift) in FL training on round $r$; \textbf{(b)} Loss landscape visualization of two local models (left) and a global model (middle), and displaying the sampling drift of different rounds (right), where the coloured blocks represent the amount of data with different classes.}
    \label{fig:fig1}
\end{figure}

Previous research has empirically and theoretically shown that data heterogeneity can degrade and destabilize the model performance in both local and global optimizations~\cite{sattler2020clustered,zhang2023no}, making the convergence slow and unstable~\cite{Wang2020Federated,nguyen2022federated,3485730}. 
Specifically, Fig.~\ref{fig:fig1} (a) provides an intuitive illustration of the local, global, and sampling drifts in FL training with \textit{non-iid} clients. 
In the local update, each client trains locally on their private dataset based on the initial model $w^{r}$ towards respective local convergence point (\textit{e.g.}, $w^{r}_{1}$ and $w^{r}_{2}$). However, local minima may not align well with the local objective (\textit{e.g.}, $w^{*}_{1}$ and $w^{*}_{2}$), thereby introducing \emph{local drift}. Additionally, the global model $w^{r+1}$ learned through aggregation at the server deviates from the optimal global model $w^{*}$. The straightforward reason for this deviation is that the client models overfit to their local convergence point, leading to an insufficient generalization of the global model (\emph{global drift}). Moreover, data heterogeneity among the subset of clients participating in each round introduces \emph{sampling drift}. This subset may exhibit a different data distribution than uniform distribution of all clients, resulting in periodic distribution shifts of the client population. As shown in Fig.~\ref{fig:fig1} (b), we further illustrate these drifts via loss landscape visualization~\cite{visualloss}, local convergence points of different clients are varied even occurring in the same round. However, since each client has its own loss landscape, the aggregated global model strays from the global objective. Additionally, the labels of different clients indicate the presence of the sampling drift.

To address such \textit{non-iid} problem in FL, existing works can be mainly divided into two directions: \romannumeral1) one direction focuses on stabilizing local training by mitigating deviation between local models and the global model in the parameter space~\cite{zhang2023fed4reid,xie2024adaptive,yang2024lightweight,li2020fedprox,tan2023federated}; \romannumeral2) another solution involves the server to enhance the effectiveness of the global weight aggregation~\cite{feddisco,zheng2023federated,fedbe,feddf}. Although these methods demonstrate a certain effect by tackling the data heterogeneity partially via either client or server updates, the gradually enlarged parameter deviation persists.

Taking into account both the effectiveness and efficiency in FL, we revisit the efforts on the underlying class-wise structure information contained in each client, which is defined as the representative embedding with identical semantics. Moreover, it represents class-relevant semantic characteristics and can be an effective carrier for communication between the server and clients. In this paper, we design a collaborative FL framework, \textbf{Fed}erated Learning with \textbf{S}tructural \textbf{K}nowledge \textbf{C}ollaboration (FedSKC), to extract and transfer the class-relevant structural knowledge across clients, effectively alleviating the detrimental effects of data heterogeneity in FL. 
Specifically, we define structural knowledge from two complementary perspectives: \textbf{\romannumeral1)} \emph{local structural knowledge}, abstracted as representative class-wise prototype features of the local model, to regulate local training by leveraging shared knowledge as an inductive bias; \textbf{\romannumeral2)} \emph{global structural knowledge}, embeds various class-relevant information collected from diverse clients, providing effective supervision signal for global aggregation and communication interaction. Notably, both local and global structural knowledge are privacy-preserving because they result from multiple averaging operations~\cite{zhu2021prototype,tan2022fedproto,tan2023federated}, and avoid the exposure of sensitive information compared to original model feature. We build local and global structural knowledge in Sec.~\ref{sec31}.

As illustrated in Fig.~\ref{fig:fig2}, FedSKC has three key components: \textbf{(\romannumeral1)} Local Contrastive Learning (LCL): FedSKC introduces an inter-contrastive learning strategy, to merge domain knowledge from different clients to regulate the local training; \textbf{(\romannumeral2)} Global Discrepancy Aggregation (GDA): FedSKC injects the discrepancy between local and global structural knowledge during the server-side aggregation, adaptively calculating the contribution weight of each client instead of relying solely on sample size; \textbf{(\romannumeral3)} Global Period Review (GPR): FedSKC combines previous and current model's parameters through the relative confidence between different structural knowledge, thus providing a more precise estimation of the model updates. 
Benefiting from these three complementary modules, FedSKC effectively unleashes the potential of class-wise structural knowledge across clients to alleviate the data heterogeneity issue, making the FL global model converge quickly and reach better performance.

Our main contributions can be summarized as follows:
\begin{itemize}
    \item We decompose data heterogeneity into local, global, and sampling drift sub-problems in FL, collectively handling these issues by extracting and transferring the structural knowledge between clients and the server.
    \item We propose a novel collaborative FL framework FedSKC, offering diverse class-relevant underlying information and stable convergence signals via structural knowledge collaboration across clients, which alleviates the challenges posed by data heterogeneity during FL training.
    \item We validate the superiority of FedSKC through extensive experiments in challenging FL scenarios (\textit{e.g.}, long-tailed, non-IIDnesses, and few-shot). Moreover, we also provide a convergence guarantee for FedSKC and derive the convergence rate under non-convex objectives (Sec.~\ref{sec33}).
\end{itemize}

\section{Related Work}
\subsection{Local Model Adjustment}
Literature on local model adjustment to improve the client-side training process, aiming at producing local models with smaller difference from the global model~\cite{li2020fedprox,moon2021,fedbe,feddf,yuan2022convergence,zhang2023fed4reid}. 
FedProx~\cite{li2020fedprox} regulate the local model training by $L_{2}$ distance between the local and global model's parameters. FedBE~\cite{fedbe} and FedDF~\cite{feddf} summarize all client models into one global model by ensemble knowledge distillation, but both rely on an unlabeled auxiliary dataset. MOON~\cite{moon2021} aligns the features on latent feature representations at the client-side to enhance the agreements between the local and global models. FedDC~\cite{gao2022federated} utilizes an auxiliary drift variable to track the gap between the local model and global model in parameter-level. All of these methods simply keep the updated local parameter close to the global model, in this way, it reduces the potential gradient divergence. However, these local methods violate the fact that the optimal point of the local empirical objective is different from global optimal point. In this work, we focus on diverse class-relevant information across clients via structural knowledge, to promote a more generalizable FL global model.

\begin{figure*}[t]
    \includegraphics[width=1.0\linewidth]{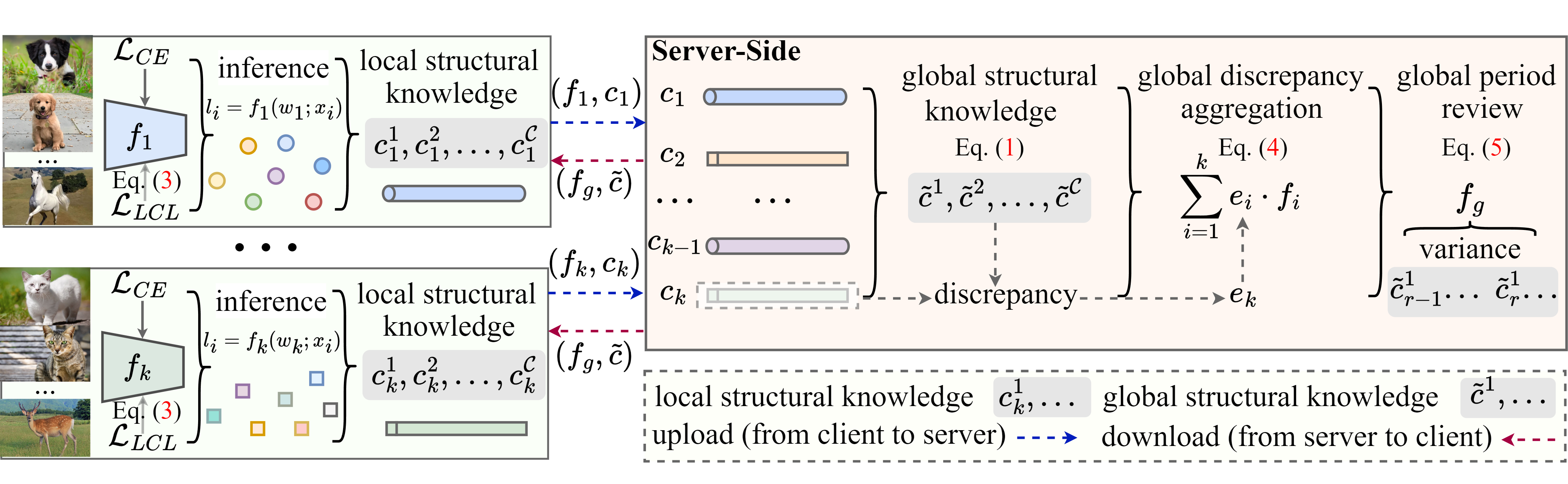}
    \caption{The overview of Federated Learning with Structural Knowledge Collaboration (FedSKC).}
    \label{fig:fig2}
\end{figure*}

\subsection{Global Model Adjustment}
Besides designs on the client-side, methods in this direction are proposed towards producing the global model with better model performance on the server-side~\cite{feddisco,fedftg,fedlc,kairouz2021advances}. 
For example, FedFTG~\cite{fedftg} explores the input space of the local model through a generator to fine-tune the global model in the server. FedLC~\cite{fedlc} effectively alleviates the global drift problem under the long-tailed distribution via logit calibration. FedMA~\cite{Wang2020Federated} constructs the shared global model in a layer-wise manner by matching and averaging the hidden elements with similar feature extraction signatures. These methods focus on aligning the local and global model to mitigate the negative impact of drifts but ignore the class-wise structure information contained in each client. FedSKC aims to jointly handle three drift sub-problems via structural knowledge collaboration.

\subsection{Contrastive Learning}
Contrastive learning has been widely applied to computer vision in the self-supervised learning scenarios~\cite{moon2021,chen2020simple,he2020momentum,le2020contrastive}. A number of works focused on constructing a positive pair and a negative pair for each instance and leveraging the InfoNCE loss~\cite{oord2018ace} to learn an encoder model, where embeddings of the same sample are pulled closer and those of different samples are pushed apart. Several works also incorporate contrastive learning into FL to assist local training~\cite{moon2021,seo2024relaxed,tan2023federated}. In this work, FedSKC builds an inter-contrastive learning strategy via structural knowledge, to regulate local model training.

\section{Methodology}
FedSKC involves three key modules to tackle local, global, and sampling drift sub-problems: Local Contrastive Learning (LCL), Global Discrepancy Aggregation (GDA), and Global Period Review (GPR). We first begin with the formulation of structural knowledge and drift sub-problems setting, then we describe the LCL, GDA, and GPR modules towards reducing these drift sub-problems. Finally, we also provide a theoretical analysis for FedSKC. The overview is shown in Fig.~\ref{fig:fig2}.

\subsection{Preliminaries}\label{sec31}
\textbf{Settings and Notations:} We first focus on a typical FL setting~\cite{li2020fedprox} with $K$ clients holding heterogeneous data partitions $\{\mathcal{D}^{1}, \mathcal{D}^{2}, \ldots, \mathcal{D}^{K}\}$ with private data $\mathcal{D}^{k}=\left\{x_{i}, y_{i}\right\}_{i=1}^{N_{k}}, y_{i}\in\{1,\ldots,\mathcal{C}\}$ for client $k$, where $N_{k}$ denotes the local data scale. Let $N_{k}^{j}$ be the number of samples with label $j$ at client $k$, and $N^{j}=\sum_{k=1}^{K}N_{k}^{j}$. In the FL training setup, for the client $k$, we consider a neural network $f_{k}$ with parameters $w_{k}=\left\{u_{k}, v_{k}\right\}$. It has two modules: a feature extractor $h_{k}(u_{k})$ with parameters $u_{k}$ maps sample $x_{i}$ to a $d$-dim feature vector $z_{i}=h_{k}(u_{k};x_{i})$; and a classifier $g_{k}(v_{k})$ with parameters $v_{k}$ maps $z_{i}$ into a $\mathcal{C}$-dim output vector as $l_{i}=g_{k}(v_{k};z_{i})$.

\textbf{Basic Federated Averaging:} With the federated averaging (FedAvg)~\cite{McMahan17}, the local objective of $k$-th client is $F_{k}$, and uses the same local epochs and learning rate. At round $r$, the server sends the initial parameters $w^{r}$ and the local model $f_{k}$ loads, while training based on dataset $\mathcal{D}^{k}$ and updating the parameter $w_{k}^{r}$. Then, the server collects and aggregates them on average $w^{r+1} \leftarrow \frac{N_{k}}{N} \sum_{k=1}^{K} w_{k}^{r}$ and then broadcasts $w^{r+1}$ in the next round, where $N$ denotes the total number of samples and $w_{k}^{r}$ denotes weights of $f_{k}$ on round $r$, repeat until convergence.

\textbf{Drift Sub-problems:} On the heterogeneous FL setting~\cite{9835537}, data heterogeneity can be described by the joint probability as $P_{i}(x,y)$=$P_{i}(x\big|y)P_{i}(y)$ between samples $x$ and labels $y$, where the $P_{i}(y)$ is different among clients. Further, we decompose it into three drift sub-problems: (\romannumeral1) \textit{local drift}: each local client has inconsistent decision boundaries and can be formulated as $f_{k_{1}}(w_{k_{1}};x) \neq f_{k_{2}}(w_{k_{2}};x)$ for the client $k_{1}$ and $k_{2}$; (\romannumeral2) \textit{global drift}: because of simple average updates on the server, global objective $F(w):=\frac{N_{k}}{N} \sum_{k=1}^{K} F_{k}(w_{k})$ can be far away from the global optima $F(w^{*})$; (\romannumeral3) \textit{sampling drift}: for the rounds $r_{1}$ and $r_{2}$, local clients are randomly selected to form subsets $s^{r_{1}}$ and $s^{r_{2}}$ from the data partition $\mathcal{D}^{1 \sim K}$, but $P_{i}(x^{r_{1}},y^{r_{1}}) \neq P_{i}(x^{r_{2}},y^{r_{2}})$, $x^{r_{1}}$ and $y^{r_{1}}$ denotes samples and labels of $s^{r_{1}}$.

\textbf{Local Structural Knowledge:} For client $k$, local structural knowledge $c_{k}^{j}$ is calculated by the normalized mean vector of features belonging to the label $j$ based on the local model $f_{k}$, which can be formulated as $c_{k}^{j}=\frac{1}{N_{k}^{j}} \sum_{i=1}^{N_{k}} f_{k}(w_{k};x_{i}\big|y_{i}=j)$, where $N_{k}^{j}$ means the number of samples with label $j$ in dataset $\mathcal{D}^{k}$. Then, we also apply a non-linear normalized transform $c_{k}^{j}=c_{k}^{j}*\text{Sigmoid}(c_{k}^{j})$ to make $c_{k}^{j}$ smoother and capture relationships among similar categories. Finally, the $k$-th client sends the group $c_{k}=\{c_{k}^{1},\ldots,c_{k}^{\mathcal{C}}\}$ to the server.

\textbf{Global Structural Knowledge:} To get the global structural knowledge $\tilde{c}^{j}$ of label $j$, local structural knowledge of clients is merged via a structure-aware adjacency matrix $\mathcal{A}^{j}$. The $\mathcal{A}^{j}$ summarizes similar local structural knowledge of label $j$ into the same group, which can be formulated as:
\begin{equation} \label{eq1}
    \mathcal{A}^{j}(k_{1}, k_{2})=\left\{
    \begin{aligned}
    1 & , \quad k_{2} \in \text{argsort}^{\mathcal{M}}_{k}\|c^{j}_{k_{1}}-c^{j}_{k}\|_2 
    \enspace\text{or}\enspace k_{2}=k_{1}; \\
    0 & , \quad \text{otherwise},
    \end{aligned}
    \right.
\end{equation}
where $\text{argsort}^{\mathcal{M}}_{k}$ denotes the top-$\mathcal{M}$ indices that would sort an array from small to large, generally setting $\mathcal{M}=1$ (the nearest client), $k\in\{1,\ldots,K\}\setminus\{k_{1}\}$ and $k_{1},k_{2}\in\{1,\ldots,K\}$. 
Then, based on Equation~\eqref{eq1}, we find similar clients of $k$-th client and merge them to get $\tilde{c}^{j}_{k}=\frac{1}{\mathcal{M}+1}\sum_{i=1}^{K}\mathcal{A}^{j}(k,i)*c^{j}_{i}$. At last, we obtain the global structural knowledge $\tilde{c}^{j}=\frac{1}{K}\sum_{k=1}^{K}\tilde{c}^{j}_{k}$ of label $j$, and then send the global structural knowledge group $\tilde{c}=\{\tilde{c}^{1},\ldots,\tilde{c}^{\mathcal{C}}\}$ to the clients participating in the FL training.

\subsection{Proposed FedSKC}\label{sec32}
\textbf{Local Contrastive Learning (LCL):} Leveraging the ideas of the contrastive learning~\cite{chen2020simple} and transfer learning~\cite{zhuang2020comprehensive}, we argue that the well-generalizable local model should provide clear decision boundaries for different labels, instead of biased boundaries limited to the local dataset. 
Furthermore, for label $j$, global structural knowledge $\tilde{c}^{j}$ can be regarded as a virtual teacher of local structural knowledge $c^{j}_{k}$ on $k$-th client. In other words, for the dataset $\mathcal{D}^{k}=\left\{x_{i}, y_{i}\right\}_{i=1}^{N_{k}}$ on $k$-th client, each sample output feature $f_{k}(w_{k};x_{i}\big|y_{i}=j)$ and global structural knowledge $\tilde{c}^{j}$ constitute a positive sample pair, $f_{k}(w_{k};x_{i}\big|y_{i}\neq j)$ and $\tilde{c}^{j}$ constitute a negative sample pair. Therefore, LCL maximizes the similarity between positive pairs and minimizes the similarity between negative pairs, to regulate local model training. Specifically, we define the similarity $s(l_{i},\tilde{c}^{j})$ of the sample output feature $l_{i}$ with global structural knowledge $\tilde{c}^{j} \in \tilde{c}=\{\tilde{c}^{1},\ldots,\tilde{c}^{\mathcal{C}}\}$, which can be formulated as:
\begin{equation} \label{eq2}
    \begin{aligned}
		& s(l_{i},\tilde{c}^{j})=\frac{1}{\mathcal{U}}(\frac{l_{i} \cdot \tilde{c}^{j}}{\left\|l_{i}\right\|_{2}\times\left\|\tilde{c}^{j}\right\|_{2}}), 
        \; \mathcal{U}=\frac{1}{N_{k}}\sum_{i=1}^{N_{k}}\left\|l_{i}-\tilde{c}^{j}\right\|_{2},
    \end{aligned}
\end{equation}
where $\mathcal{U}$ is a normalization term for similarity, $l_{i}=f_{k}(w_{k};x_{i})$ and we set $\mathcal{U}$ denotes the average distance between $l_{i}|_{i=1}^{N_{k}}$ and $\tilde{c}^{j}$ on dataset $\mathcal{D}^{k}$. Thus, based on Equation \eqref{eq2}, we can measure the similarity among different sample pairs. Then, we devise local contrastive learning (LCL) to quantify and regulate local training, LCL can be formulated as the following objective:
\begin{equation} \label{eq3}
    \begin{aligned}
    \mathcal{L}_{LCL} & = -\text{log}\frac{\text{exp}(s(l_{i},\tilde{c}^{y_{i}})/\tau)}{\sum_{j=1}^{\mathcal{C}}\text{exp}(s(l_{i},\tilde{c}^{j})/\tau)} \\
    & = \text{log}(1 + \frac{\sum_{\tilde{c}^{j} \in \tilde{c}\setminus\{\tilde{c}^{y_{i}}\}}\text{exp}(s(l_{i},\tilde{c}^{j})/\tau)}{\text{exp}(s(l_{i},\tilde{c}^{y_{i}})/\tau)}),
    \end{aligned}
\end{equation}
where $\tau$ is a temperature parameter to control the representation strength~\cite{chen2020simple}. For the $k$-th client, note that by minimizing Equation \eqref{eq3}, local model $f_{k}$ brings each sample output feature $l_{i}$ close to the representation of truth label $y_{i}$ and away from the representation of other false labels. In other words, global structural knowledge $\tilde{c}^{j}$ can be regarded as a virtual teacher to regulate local model training on the client's private dataset, providing a clear class-wise decision boundary for label $j$.

\textbf{Global Discrepancy Aggregation (GDA):} The FedAvg~\cite{McMahan17} global aggregation method may be far from optimal because the dataset size does not reflect any client-specific information (\textit{e.g.}, category distribution). Instead, we consider the discrepancy between local structural knowledge and global structural knowledge to optimize the aggregation weight of each client, and design the global discrepancy aggregation (GDA) strategy. Specifically, client $k$ can calculate the discrepancy $d_{k}$ between its local structural knowledge and the global structural knowledge, which can be formulated as $d_{k}=\sum_{j=1}^{\mathcal{C}}\|c_{k}^{j}-\tilde{c}^{j}\|_{2}$. We speculate that client $k$ with larger dataset size $N_{k}$ and smaller discrepancy $d_{k}$ would have a higher aggregation weight $e_{k}$:
\begin{equation} \label{eq4}
    e_{k} = \frac{\text{Sigmoid}(N_{k} - a_{k} \cdot d_{k} + b_{k})}
    {\sum_{i=1}^{K}\text{Sigmoid}(N_{i} - a_{i} \cdot d_{i} + b_{i})},
\end{equation}
where $a_{k}=d_{k} / \sum_{i=1}^{K}d_{i}$ and $b_{k}=N_{k} / \sum_{i=1}^{K}N_{i}$ are used to balance $d_{k}$ and $N_{k}$. Intuitively, the $d_{k}$ can be regarded as a proximal term to alleviate the negative impacts of dataset size in global aggregation. Therefore, the parameters of the global model $f_{g}(w_{g})$ can be computed as $w_{g}=\sum_{k=1}^{K}e_{k} \cdot w_{k}$.

\textbf{Global Period Review (GPR):} Although incorporating the discrepancy between the local and global structural knowledge into global aggregation could be beneficial, the optimization objective might deviate from the overall aim of all clients due to sampling drift, which would make the global model difficult to converge. Therefore, based on momentum optimization, we would assume the previous global structural knowledge could regulate current global model updates. GPR combines global model $f_{g}^{r-1}(w_{g}^{r-1})$ and $f_{g}^{r}(w_{g}^{r})$ on round $r-1$ and $r$ via linear interpolation. The coefficient indicates the relative confidence between $f_{g}^{r-1}$ and $f_{g}^{r}$, which is determined by the variance of global structural knowledge group $\tilde{c}_{r-1}$ and $\tilde{c}_{r}$, thus producing more stable and accurate model updates. For global structural knowledge $\tilde{c}^{j}_{r}$ of label $j$ on round $r$, the variance of $\tilde{c}^{j}_{r}$ can be calculated as $\sigma_{r(j)}^{2}=\frac{1}{\mathcal{C}}\sum_{i=1}^{\mathcal{C}}(\tilde{c}^{j(i)}_{r}-\mu(\tilde{c}^{j}_{r}))^{2}$, where $\tilde{c}^{j(i)}_{r}$ is the $i$-th value of $\tilde{c}^{j}_{r}$ and $\mu(\tilde{c}^{j}_{r})=\frac{1}{\mathcal{C}}\sum_{i=1}^{\mathcal{C}}\tilde{c}^{j(i)}_{r}$. GPR combines $w_{g}^{r-1}$ and $w_{g}^{r}$ based on the discrepancy in variances:
\begin{equation} \label{eq5}
    w_{g}^{r} = \beta \cdot w_{g}^{r}+(1-\beta)\cdot\frac{\sum_{j=1}^{\mathcal{C}}\big(\sigma_{r(j)}^{2}-\sigma_{r-1(j)}^{2}\big)}
    {\sum_{j=1}^{\mathcal{C}}\sigma_{r-1(j)}^{2}}(w_{g}^{r-1} - w_{g}^{r}),
\end{equation}
where $\beta$ denotes the momentum hyper-parameter to balance $w_{g}^{r-1}$ and $w_{g}^{r}$, generally setting $\beta=0.95$. Based on Equation \eqref{eq5}, we expect to review and adjust the update direction of the global model through the relative relationship between $\sigma_{r-1(j)}^{2}$ and $\sigma_{r(j)}^{2}$. Finally, updated parameters $w_{g}^{r}$ of the global model $f_{g}^{r}(w_{g}^{r})$ are sent to clients for the next round of FL training.

\subsection{Theoretical Analysis} \label{sec33}
To theoretically analyze for FedSKC, we denote the objective function $\mathcal{L}_{LCL}+\mathcal{L}_{CE}$ to be defined as $\mathcal{L}_{r}$ with a subscript indicating the round $r$ of iterations. This analysis is based on four standard assumptions (Appendix), which are commonly used in different FL literature~\cite{tan2022fedproto,li2020fedprox,feddisco}. Here, we present the theoretical results of FedSKC under non-convex objectives:
\begin{theorem}[Deviation bound of the objective function] \label{theorem1}
    With the assumptions, FedSKC loss function $\mathcal{L}$ of an arbitrary client will be bounded as follows:
    \begin{equation}
        \begin{aligned}
        \mathbb{E}[\mathcal{L}_{(r+1)E}] & \leq \mathcal{L}_{rE}-\left(\eta-\frac{L_{1}\eta^{2}}{2}\right)EB^{2} \\
        & \quad + \frac{L_{1}E\eta^{2}}{2}\sigma^{2}+
        \frac{L_{2}E\eta|\mathcal{C}|B}{|\mathcal{M}|+1},
        \end{aligned}
    \end{equation}
\end{theorem}
we express $\mathbf{e} \in \{1,2,\ldots,E\}$ as local iteration step, $r$ as the global round, $rE+\mathbf{e}$ refers to the $\mathbf{e}$-th local update in round $r+1$, $rE+1$ denotes the time between $r$-th global aggregation at the server and starting the first update on the local model, $\eta$ is the local learning rate, $\{\mathcal{C},\mathcal{M}\}$ are constants in Sec.~\ref{sec31}, $\{L_{1},L_{2},B,\sigma^{2}\}$ are constants in assumptions from Appendix. Theorem~\ref{theorem1} indicates that $\mathcal{L}_{r}$ decreases as round $r$ increases, convergence can be guaranteed by choosing an appropriate $\eta$.
\begin{theorem}[Non-convex convergence of the FedSKC] \label{theorem2}
    With the assumptions, FedSKC loss function $\mathcal{L}$ of an arbitrary client decreases monotonically as the round increases, when:
    \begin{equation}
        \begin{aligned}
        \eta_{\mathbf{e}} < \frac{2(|\mathcal{M}|+1)B^{2}-2L_{2}|\mathcal{C}|B}{L_{1}(|\mathcal{M}|+1)(\sigma^{2}+B^{2})} \enspace 
        \textnormal{where} \enspace \mathbf{e} \in \{1,\ldots,E-1\}.
        \end{aligned}
    \end{equation}
\end{theorem}
Thus, based on Theorem~\ref{theorem1} and Theorem~\ref{theorem2}, the loss function $\mathcal{L}$ to be negative, which can guide the choice of appropriate values for the learning rate $\eta$ to guarantee the convergence.
\begin{theorem}[Non-convex convergence rate of the FedSKC]
    \label{theorem3}
    With the assumptions, let the global round $r$ from $0$ to $R-1$, given any $\xi>0$, 
    $\mathcal{L}$ will converge, when:
    \begin{equation}
        \begin{aligned}
        & R > \frac{2(|\mathcal{M}|+1)(\mathcal{L}_{0}-\mathcal{L}^{*})}{\xi E\eta(|\mathcal{M}|+1)(2-L_{1}\eta)-\mathcal{P}-\mathcal{H}} \\
        & \textnormal{where} \enspace {\eta} < \frac{2 \xi (|\mathcal{M}|+1) - 2 L_{2}|\mathcal{C}|B}
                    {L_{1}(|\mathcal{M}|+1)(\xi + \sigma^{2})},
        \end{aligned}
    \end{equation}
\end{theorem}
where $\mathcal{P}=(|\mathcal{M}|+1)L_{1}E\eta^{2}\sigma^{2}$, $\mathcal{H}=2L_{2}E\eta|\mathcal{C}|B$, and $\mathcal{L}^{*}$ denotes the optimal solution. \textit{Please note that all assumptions and proofs are formally given in the Appendix~\ref{app:ana}}.

\renewcommand{\algorithmicrequire}{\textbf{RunServer}($f_{k}, c_{k}$):}
\renewcommand{\algorithmicensure}{\textbf{RunClient}($f_{g}, \tilde{c}$):}
\begin{algorithm}[!htb]
    \caption{Pseudo-code Flow of FedSKC.}
    \label{alg:fedskc}
    \begin{algorithmic}[1]
        \Require \emph{\textcolor[RGB]{0,0,255}{/* \quad Global Server \quad */}} 
        \State Initialized rounds $R$, number of clients $K$ and labels $\mathcal{C}$, global model $f_{g}$, and global structural knowledge $\tilde{c}$
        \For{each round $r = 1,2,...,R$}
            \State $A_{r} \leftarrow$ (server selects a random subset of $K$ clients)
            \For{each client $ k \in A_{r} $ \textbf{in parallel}}
                \State $f_{k}, c_{k} \leftarrow \text{RunClient}(f_{g}, \tilde{c})$
            \EndFor
            \State $\mathcal{A} \leftarrow$ (structure-aware adjacency matrix by Eq.~\eqref{eq1})
            \For{each label $j = 1,2,...,\mathcal{C}$}
                \For{each client $ k \in A_{r} $ \textbf{in parallel}}
                    \State $\tilde{c}^{j}_{k}=\frac{1}{\mathcal{M}+1}\sum_{i=1}^{|A_{r}|}\mathcal{A}^{j}(k,i)*c^{j}_{i}$ 
                \EndFor
                \State $\tilde{c}^{j}=\frac{1}{|A_{r}|}\sum_{k \in A_{r}}\tilde{c}^{j}_{k}$
            \EndFor
            \State $\tilde{c}=\{\tilde{c}^{1},\ldots,\tilde{c}^{\mathcal{C}}\}$ \emph{\textcolor[RGB]{0,0,255}{// global structural knowledge $\tilde{c}$}}
            \State \emph{\textcolor[RGB]{0,0,255}{// Global Discrepancy Aggregation (GDA)}} 
            \For{each client $ k \in A_{r} $ \textbf{in parallel}}
                \State $d_{k}=\sum_{j=1}^{\mathcal{C}}\|c_{k}^{j}-\tilde{c}^{j}\|_{2}$ 
                \State $e_{k} \leftarrow$ (aggregation weight of client $k$ by Eq.~\eqref{eq4})
            \EndFor
            \State $f_{g}=\sum_{k \in A_{r}}e_{k} \cdot f_{k}$ 
            \State \emph{\textcolor[RGB]{0,0,255}{// Global Period Review (GPR)}}
            \State $f_{g}^{r-1}, \tilde{c}_{r-1} \leftarrow$ (for the previous round $r-1$)
            \For{each label $j = 1,2,...,\mathcal{C}$}
                \State $\mu(\tilde{c}^{j})=\frac{1}{\mathcal{C}}\sum_{i=1}^{\mathcal{C}}\tilde{c}^{j(i)}$ \emph{\textcolor[RGB]{0,0,255}{// mean of $\tilde{c}$}}
                \State $\sigma_{(j)}^{2}=\frac{1}{\mathcal{C}}\sum_{i=1}^{\mathcal{C}}(\tilde{c}^{j(i)}-\mu(\tilde{c}^{j}))^{2}$ \emph{\textcolor[RGB]{0,0,255}{// variance of $\tilde{c}$}}
            \EndFor
            \State $f_{g} \leftarrow$ (adjust the global model by Eq.~\eqref{eq5})
        \EndFor
        \Ensure \emph{\textcolor[RGB]{0,0,255}{/* \quad Local Client \quad */}} 
        \State Initialized local epochs $E$, data size $N_{k}$, learning rate $\eta$
        \For{each local epoch $\mathbf{e} = 1,2,...,E$} 
            \State \emph{\textcolor[RGB]{0,0,255}{// Local Contrastive Learning (LCL)}}
            \For{each batch $b \in$ client dataset $\mathcal{D}^{k}$}
                \State $\mathcal{L}_{CE}=\sum_{i\in b}-y_{i}\textnormal{log}(\textnormal{softmax}(f_{k}(x_{i})))$ 
                \State $\mathcal{L}_{LCL} \leftarrow$ ($l_{i},\tilde{c},\tilde{c}^{y_{i}}$) by Eq.~\eqref{eq2} and Eq.~\eqref{eq3}
                \State $\ell = \mathcal{L}_{CE} + \mathcal{L}_{LCL}$
                \State $f_{k} \leftarrow f_{k}-\eta \nabla \ell(f_{k}; b)$ \emph{\textcolor[RGB]{0,0,255}{// model update by SGD}}
            \EndFor	
        \EndFor
        \State $c_{k}=\{c_{k}^{1},\ldots,c_{k}^{\mathcal{C}}\}$ \emph{\textcolor[RGB]{0,0,255}{// local structural knowledge $c_{k}$}}
        \State Return $f_{k}$ and $c_{k}$ to the server.
    \end{algorithmic}
\end{algorithm}

\begin{table*}[t]
	\caption{The Top-1 test accuracy ($\%$) for FedSKC and compared baselines on CIFAR-10, CIFAR-100, CIFAR-10-LT, and FC100.}
	\label{tab:tab1}
	\centering
        \resizebox{\linewidth}{!}{
    	\begin{tabular}{lccccccccc}
    		\toprule[1.0pt]
    		\multirow{3}{*}{\textbf{FL Baselines}} & \multicolumn{2}{c}{\textbf{CIFAR-10}}& \multicolumn{2}{c}{\textbf{CIFAR-100}}
    		& \multicolumn{3}{c}{\textbf{CIFAR-10-LT} ($\alpha=0.2$)} & \multicolumn{2}{c}{\textbf{FC100} (5-way 5-shot)} \\
    		\cmidrule(lr){2-3} \cmidrule(lr){4-5} \cmidrule(lr){6-8} \cmidrule(lr){9-10}
    		& $\alpha=0.05$ & $\alpha=0.2$ 
                & $\alpha=0.05$ & $\alpha=0.2$ 
                & $\rho=100$ & $\rho=50$ & $\rho=10$ 
    		& $\alpha=0.05$ & $\alpha=0.2$ \\
    		\cmidrule(lr){1-1} \cmidrule(lr){2-3} \cmidrule(lr){4-5} \cmidrule(lr){6-8} \cmidrule(lr){9-10}
                $\text{FedAvg}_{[\text{AISTAT'17}]}$ & 73.55 &78.61 & 52.47 & 55.67 & 53.09 & 54.56 & 62.78 & 47.40 & 50.24 \\
                $\text{FedProx}_{[\text{MLSys'20}]}$ & 73.06 & 78.90 & 52.13 & 55.28 & 53.52 & 54.36 & 62.12 & 47.06 & 50.43 \\
                $\text{Scaffold}_{[\text{ICML'20}]}$ & 73.48 & 78.25 & 52.27 & 55.30 & 53.10 & 55.60 & 63.19 & 47.54 & 50.35 \\
                $\text{FedOPT}_{[\text{ICLR'21}]}$ & 73.23 & 78.80 & 52.62 & 55.15 & 53.47 & 54.71 & 63.11 & 47.31 & 50.60 \\
                $\text{MOON}_{[\text{CVPR'21}]}$ & 75.79 & 81.14 & 55.04 & 57.26 & 54.34 & 57.88 & 64.03 & 48.25 & 50.81 \\
                $\text{FedProto}_{[\text{AAAI'22}]}$ & 74.69 & 79.72 & 53.67 & 56.03 & 54.12 & 56.29 & 63.95 & 45.76 & 48.92 \\
                $\text{FedNH}_{[\text{AAAI'23}]}$ & 75.02 & 80.66 & 54.23 & 56.89 & \underline{56.32} & \underline{58.75} & 64.36 & 48.68 & \underline{51.22} \\
                $\text{FedRCL}_{[\text{CVPR'24}]}$ & \underline{76.14} & \underline{81.30} & \underline{55.34} & \underline{57.31} & 56.09 & 58.39 & \underline{64.60} & \underline{48.82} & 51.13 \\
    		\midrule[0.66pt]
    		\rowcolor{myblue}
    		FedSKC  & \textbf{77.13} & \textbf{82.35} & \textbf{56.62} & \textbf{58.15} & \textbf{57.55} & \textbf{60.04} & \textbf{65.72} & \textbf{49.65} & \textbf{51.80} \\
    		\bottomrule[1.0pt]
    	\end{tabular}
        }
\end{table*}

\subsection{Discussion}
\textbf{Summary:} In each communication round, server distributes global structural knowledge group $\tilde{c}$ to participants, and aggregates and adjusts the global model based on GDA and GPR. In local training, each participant optimizes on the local private data, while the objective function as $\mathcal{L}=\mathcal{L}_{LCL}+\mathcal{L}_{CE}$, where $\mathcal{L}_{CE}=-y_{i}\text{log}(\text{softmax}(f(w;x_{i})))$ is CrossEntropy~\cite{celoss2005} loss, then each client $k$ sends local structural knowledge group $c_{k}$ to the server. After iterative updates, each local model gradually approaches an optimal solution, promoting the aggregation and improvement of the global model until reaching a convergence. The pseudo-code flow of FedSKC is shown in Algorithm~\ref{alg:fedskc}.

\textbf{Structural Knowledge:} The conventional prototype learning paradigm~\cite{pl2018robust,protonet2017} calculates the mean of feature vectors within every label as a prototype, but the prototype might lead to ambiguities due to the FL-specific data heterogeneity issue. Thus, we expect to extract more representative local structural knowledge of each client's local model, while maintaining the more equilibrial global structural knowledge on the server. \textit{On the one hand}, local structural knowledge is abstracted into the normalized pre-softmax output of the local model, which can generate highly diverse class-relevant knowledge at the local-level. \textit{On the other hand}, global structural knowledge can be regarded as mutually invariant features among heterogeneous clients, which can be considered as a fair teacher to provide an unbiased convergence objective at the global-level.

\textbf{Privacy:} FedSKC is privacy-preserving because the structural knowledge is transferred between the server and clients without leaking the client's exact distribution. Especially, calculating local structural knowledge requires extracting mean representation vectors of samples and performing a non-linear normalized transform, which is an irreversible operation. The server only collects the discrepancy between local and global structural knowledge for global aggregation, compared to the basic FedAvg~\cite{McMahan17}, further enhancing the protection of privacy. Besides, due to multiple operations, it is less feasible to reconstruct private data backward from the structural knowledge.

\section{Experiments}
\subsection{Experimental Setup}
\textbf{Datasets and Models:} Based on different FL scenarios, we conduct experiments on: \romannumeral1) \textit{non-IIDnesses}: we use three image classification datasets (EMNIST~\cite{emnist2017cohen}, CIFAR-10 and CIFAR-100~\cite{cifar2009krizhevsky}). We use a multi-layers CNN for EMNIST, ResNet-8~\cite{resnet8ju} for both CIFAR-10 and CIFAR-100; \romannumeral2) \textit{long-tailed}: we shape the original CIFAR-10 and CIFAR-100 into long-tailed version with $\rho=10,50,100$ follow~\cite{caoKaidi2019} to get CIFAR-10-LT and CIFAR-100-LT, and same models as above; \romannumeral3) \textit{few-shot}: FC100~\cite{fc1002018tadam} is a split dataset based on CIFAR-100 for few-shot classification and we use ProtoNet~\cite{protonet2017} for training.

\textbf{Configurations:} Unless otherwise mentioned, we run $R=200$ global rounds, $K=20$ clients in total, the active ratio $\epsilon=0.4$, \textit{non-iid} ratio $\alpha=0.2$ follows Dirichlet distribution~\cite{Wang2020Federated}. We adopt a local epoch $E=10$ and each step uses a mini-batch with size $64$, and we set $\mathcal{M}=1$ in Eq.~\eqref{eq1}, $\tau=0.08$ in Eq.~\eqref{eq3}, $\beta=0.95$ in Eq.~\eqref{eq5}. We unify hyperparameter settings for all clients and use SGD with the initial learning rate of $0.01$ as the optimizer to update model parameters. Experiments are run by PyTorch on two NVIDIA GeForce RTX 3060 GPUs.

\textbf{Baselines:} We compare our FedSKC with several SOTA FL methods: standard (FedAvg~\cite{McMahan17}, FedProx~\cite{li2020fedprox}); contrast-based (MOON~\cite{moon2021}, FedRCL~\cite{seo2024relaxed}); prototype-based (FedProto~\cite{tan2022fedproto}, FedNH~\cite{fednh2023}); aggregation-based (Scaffold~\cite{pmlr-v119-karimireddy20a}, FedOPT~\cite{fedopt21}). Note that all results reported are the average of three repeating runs with a standard deviation of different random seeds.

\begin{table}[t]
	\caption{The Top-1 test accuracy ($\%$) for FedSKC and compared baselines on EMNIST (split by merge, 47 labels) and CIFAR-100-LT datasets.}
	\label{tab:tab11}
	\centering
    \resizebox{\linewidth}{!}{
    	\begin{tabular}{lccccc}
    		\toprule[1.0pt]
    		\multirow{3}{*}{\textbf{Baselines}} &\multicolumn{2}{c}{\textbf{EMNIST}} & \multicolumn{3}{c}{\textbf{CIFAR-100-LT} ($\alpha=0.2$)} \\
    		\cmidrule(lr){2-3} \cmidrule(lr){4-6}
    		& $\alpha=0.05$ & $\alpha=0.2$ & $\rho=100$ & $\rho=50$ & $\rho=10$ \\
    		\cmidrule(lr){1-1} \cmidrule(lr){2-3} \cmidrule(lr){4-6} 
                FedAvg & 80.17 & 84.76 & 30.37 & 32.67 & 42.33 \\
                FedProx & 80.03 & 84.60 & 30.41 & 32.30 & 42.60 \\
                Scaffold & 80.15 & 84.74 & 30.47 & 32.22 & 42.35 \\
                FedOPT & 80.69 & 84.55 & 30.10 & 32.40 & 42.23 \\
                MOON & 82.33 & 87.13 & 31.89 & 33.41 & 43.57 \\
                FedProto & 81.53 & 85.29 & 30.85 & 32.95 & 43.03 \\
                FedNH & 81.79 & 86.33 & \underline{32.33} & \underline{34.27} & 43.89 \\
                FedRCL & \underline{82.63} & \underline{87.15} & 32.28 & 33.97 & \underline{44.18} \\
    		\midrule[0.66pt]
    		\rowcolor{myblue}
    		FedSKC  &\textbf{83.75} &\textbf{88.16} &\textbf{33.20} &\textbf{35.19} &\textbf{45.08} \\
    		\bottomrule[1.0pt]
    	\end{tabular}
    }
\end{table}

\subsection{Evaluation Results}
\textbf{Performance over Baselines:} The comparison results with CIFAR-10, CIFAR-100, CIFAR-10-LT, and FC100 for different heterogeneity levels $\alpha$ and imbalanced levels $\rho$ are shown in Tab.~\ref{tab:tab1}. We can observe that FedSKC universally outperforms baselines, which implies that FedSKC can effectively utilize structural knowledge to promote model training. Experiments show that: \textbf{\romannumeral1)} As the data heterogeneity increases ($\alpha$ decreases), FedSKC consistently outperforms baselines, indicating clearly that FedSKC is robust to different heterogeneity levels $\alpha$; \textbf{\romannumeral2)} FedSKC achieves significantly better on CIFAR-10-LT setting ($\rho=100$, compared with FedAvg gains reach 4.46\%), to show the effectiveness of FedSKC under the more difficult task for severe heterogeneity and imbalance, which also confirms the GDA module in Eq.~\eqref{eq4} helps alleviate long-tailed distribution; \textbf{\romannumeral3)} For the few-shot setting, FedSKC potentially enhances the variety of sample features by transferring the global structural knowledge, achieving better results than baselines. Results on EMNIST and CIFAR-100-LT are given in Tab.~\ref{tab:tab11}.

\begin{table*}[t]
	\caption{The comparison results of convergence rates for FedSKC and compared baselines on CIFAR-10, CIFAR-100, and CIFAR-10-LT.}
	\label{tab:tab2}
	\centering
        \resizebox{\linewidth}{!}{
    	\begin{tabular}{lcccccccccc}
    		\toprule[1.0pt]
    		\multirow{3}{*}{\textbf{FL Baselines}} & \multicolumn{4}{c}{\textbf{CIFAR-10} ($\alpha=0.2$)}& \multicolumn{3}{c}{\textbf{CIFAR-100} ($\alpha=0.2$)}
    		& \multicolumn{3}{c}{\textbf{CIFAR-10-LT} ($\rho=50$)} \\
    		\cmidrule(lr){2-5} \cmidrule(lr){6-8} \cmidrule(lr){9-11}
    		& $R_{20\%}$ & $R_{40\%}$ & $R_{75\%}$ & $Acc_{[\%]}$ 
                & $R_{20\%}$ & $R_{50\%}$ & $Acc_{[\%]}$ 
                & $R_{20\%}$ & $R_{50\%}$ & $Acc_{[\%]}$  \\
    		\cmidrule(lr){1-1} \cmidrule(lr){2-5} \cmidrule(lr){6-8} \cmidrule(lr){9-11}
                $\text{FedAvg}_{[\text{AISTAT'17}]}$ & 6 & 23 & 165 & 78.61 & 18 & 105 & 55.67 & 6 & 108 & 54.56 \\
                $\text{FedProx}_{[\text{MLSys'20}]}$ & 6 & 20 & 162 & 78.90 & 18 & 98 & 55.28 & 8 & 103 & 54.36 \\
                $\text{Scaffold}_{[\text{ICML'20}]}$ & 8 & 27 & 170 & 78.25 & 25 & 112 & 55.30 & 8 & 116 & 55.60 \\
                $\text{FedOPT}_{[\text{ICLR'21}]}$ & 7 & 30 & 165 & 78.80 & 25 & 115 & 55.15 & 10 & 122 & 54.71 \\
                $\text{MOON}_{[\text{CVPR'21}]}$ & 5 & 13 & 108 & 81.14 & 13 & 76 & 57.26 & 6 & 85 & 57.88 \\
                $\text{FedProto}_{[\text{AAAI'22}]}$ & 6 & 20 & 155 & 79.72 & 18 & 94 & 56.03 & 6 & 95 & 56.29 \\
                $\text{FedNH}_{[\text{AAAI'23}]}$ & 6 & 16 & 119 & 80.66 & 18 & 83 & 56.89 & 5 & 77 & \underline{58.75} \\
                $\text{FedRCL}_{[\text{CVPR'24}]}$ & 5 & 11 & 97 & \underline{81.30} & 10 & 70 & \underline{57.31} & 5 & 75 & 58.39 \\
    		\midrule[0.66pt]
    		\rowcolor{myblue}
    		FedSKC  &\textbf{3}&\textbf{8}&\textbf{84}&\textbf{82.35} 
                        &\textbf{9}&\textbf{64}&\textbf{58.15} 
                        &\textbf{3}&\textbf{61}&\textbf{60.04} \\
    		\bottomrule[1.0pt]
    	\end{tabular}
        }
\end{table*}

\begin{table}[t]
	\caption{The Comparison results of convergence rates on EMNIST (split by merge, 47 labels) and CIFAR-100-LT datasets.}
	\label{tab:tab22}
	\centering
    \resizebox{\linewidth}{!}{
    	\begin{tabular}{lcccccc}
    		\toprule[1.0pt]
    		\multirow{3}{*}{\textbf{Baselines}} &\multicolumn{3}{c}{\textbf{EMNIST ($\alpha=0.2$)}} &\multicolumn{3}{c}{\textbf{CIFAR-100-LT} ($\rho=50$)} \\
    		\cmidrule(lr){2-4} \cmidrule(lr){5-7}
    		& $R_{70\%}$ & $R_{80\%}$ & $Acc_{[\%]}$ & $R_{15\%}$ & $R_{30\%}$ & $Acc_{[\%]}$  \\
    		\cmidrule(lr){1-1} \cmidrule(lr){2-4} \cmidrule(lr){5-7}
                FedAvg & 51 & 116 & 84.76 & 24 & 108 &32.67 \\
                FedProx & 46 & 112 & 84.60 & 21 & 97 & 32.30 \\
                Scaffold & 53 & 120 & 84.74 & 30 & 115 & 32.22 \\
                FedOPT & 50 & 123 & 84.55 & 25 & 113 & 32.40 \\
                MOON & 36 & 75 & 87.13 & 18 & 85 & 33.41 \\
                FedProto & 46 & 107 & 85.29 & 21 & 92 & 32.95 \\
                FedNH & 42 & 96 & 86.33 & 17 & 79 & \underline{34.27} \\
                FedRCL & 36 & 71 & \underline{87.15} & 17 & 83 & 33.97 \\
    		\midrule[0.66pt]
    		\rowcolor{myblue}
    		FedSKC & \textbf{28} & \textbf{63} & \textbf{88.16} & \textbf{13} & \textbf{68} & \textbf{35.19} \\
    		\bottomrule[1.0pt]
    	\end{tabular}
    }
\end{table}

\textbf{Convergence Rates:} The comparison results on CIFAR-10, CIFAR-100, CIFAR-10-LT are shown in Tab.~\ref{tab:tab2}, where the $R_{acc}$ columns denote the minimum number of rounds required to reach $acc$ of the test accuracy, and the $Acc_{[\%]}$ columns show the final accuracy. Experiments show that: \textbf{\romannumeral1)} FedSKC required fewer rounds to converge compared with other baselines and achieved higher accuracy, showing the superiority of FedSKC in dealing with sampling drift; \textbf{\romannumeral2)} FedProx converges slightly faster than FedAvg, because it alleviates the deviation of the optimization direction via the proximal term. Similarly, LCL module in FedSKC via Eq.~\eqref{eq3} enhances the discriminability in the feature space, thereby accelerating convergence. Results on EMNIST and CIFAR-100-LT are given in Tab.~\ref{tab:tab22}.

\begin{figure*}[t]
    \centering
    \begin{minipage}[t]{0.249\linewidth}
        \centering
        \subfigure[Temperature $\tau$]{\label{fig:fig3a}\includegraphics[width=1.0\linewidth]{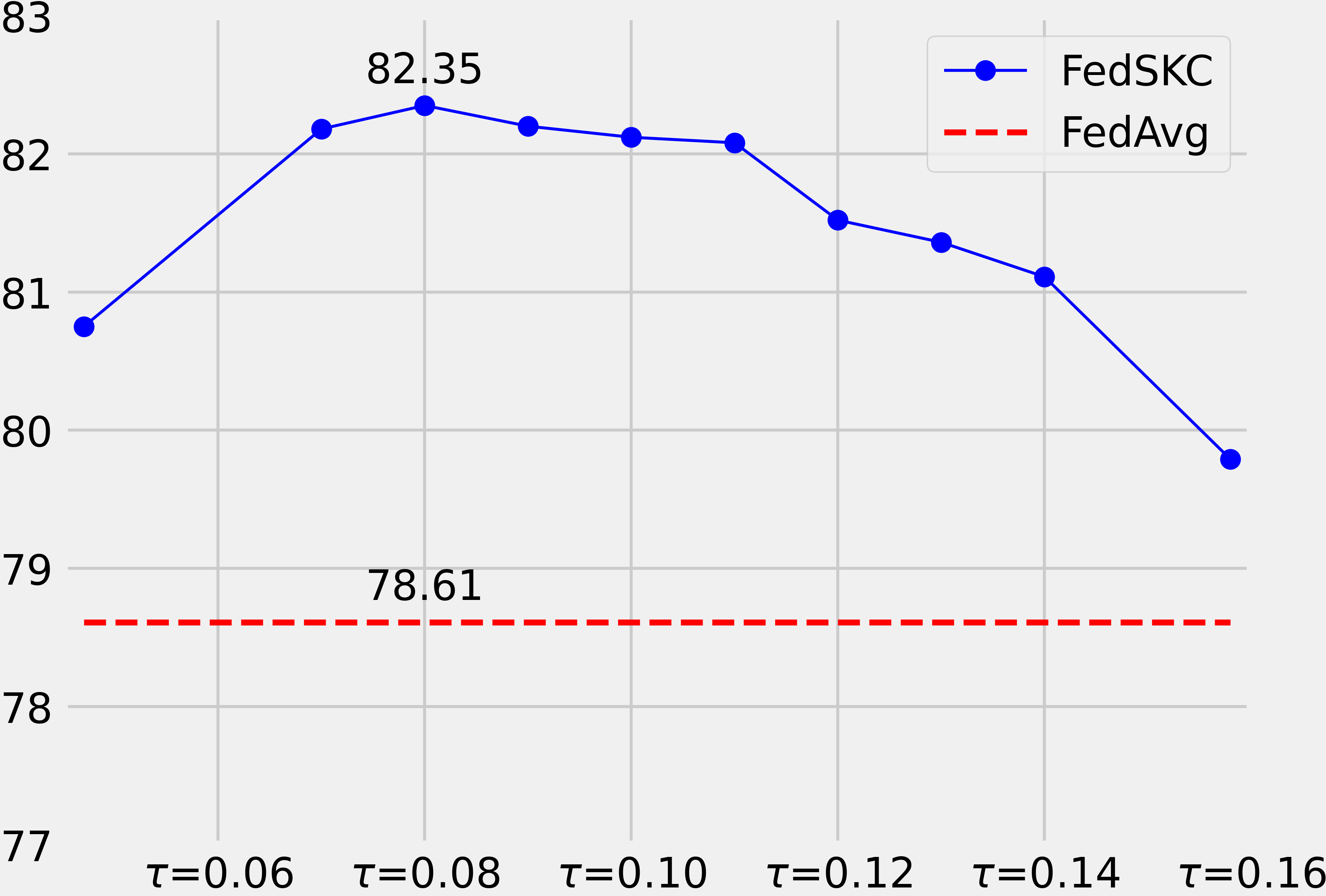}}
    \end{minipage}
    \begin{minipage}[t]{0.241\linewidth}
        \centering
        \subfigure[Similar Clients $\mathcal{M}$]{\label{fig:fig3b}\includegraphics[width=0.98\linewidth]{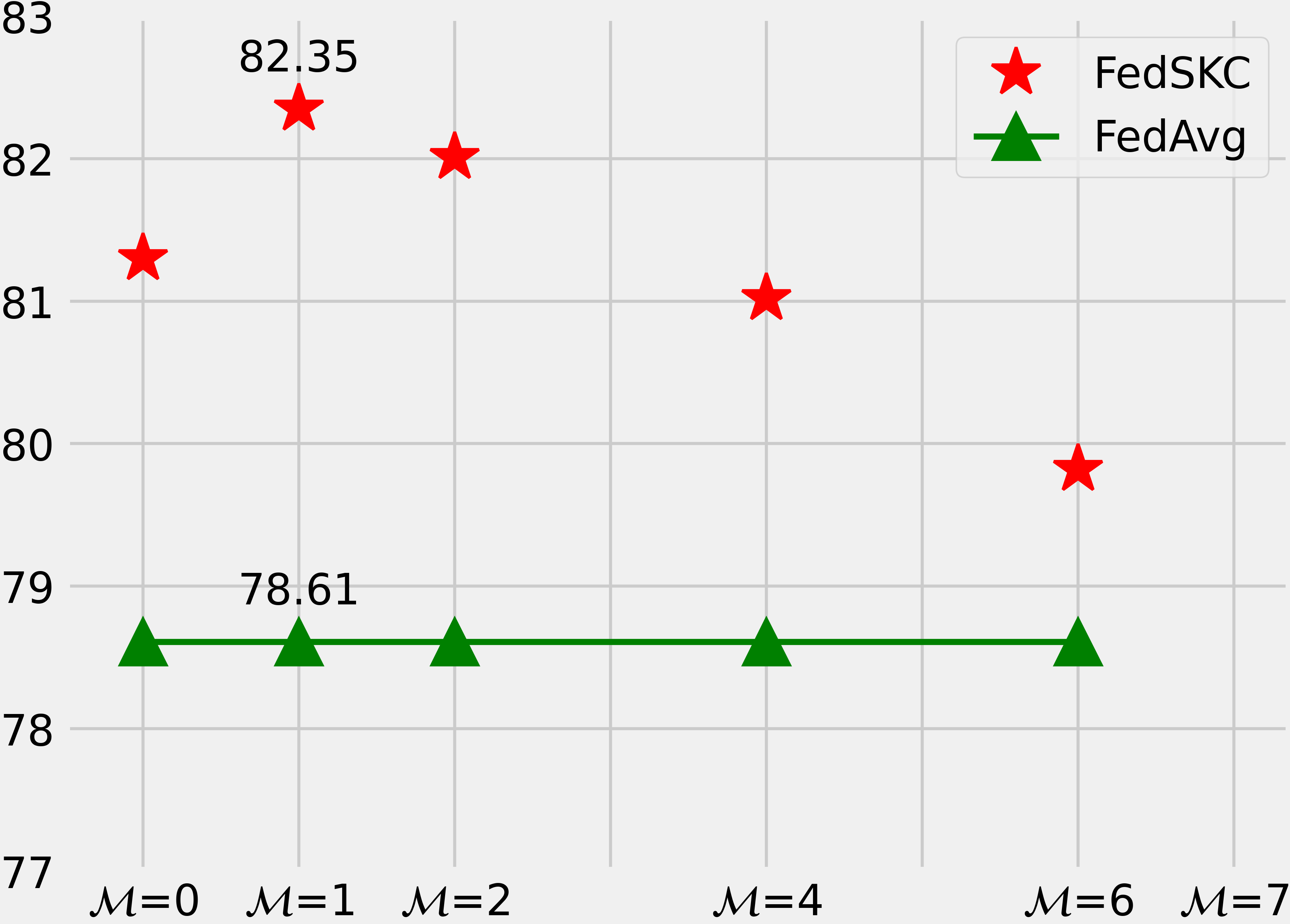}}
    \end{minipage}
    \begin{minipage}[t]{0.246\linewidth}
        \centering
        \subfigure[Heterogeneity Level $\alpha$]{\label{fig:fig3c}\includegraphics[width=0.98\linewidth]{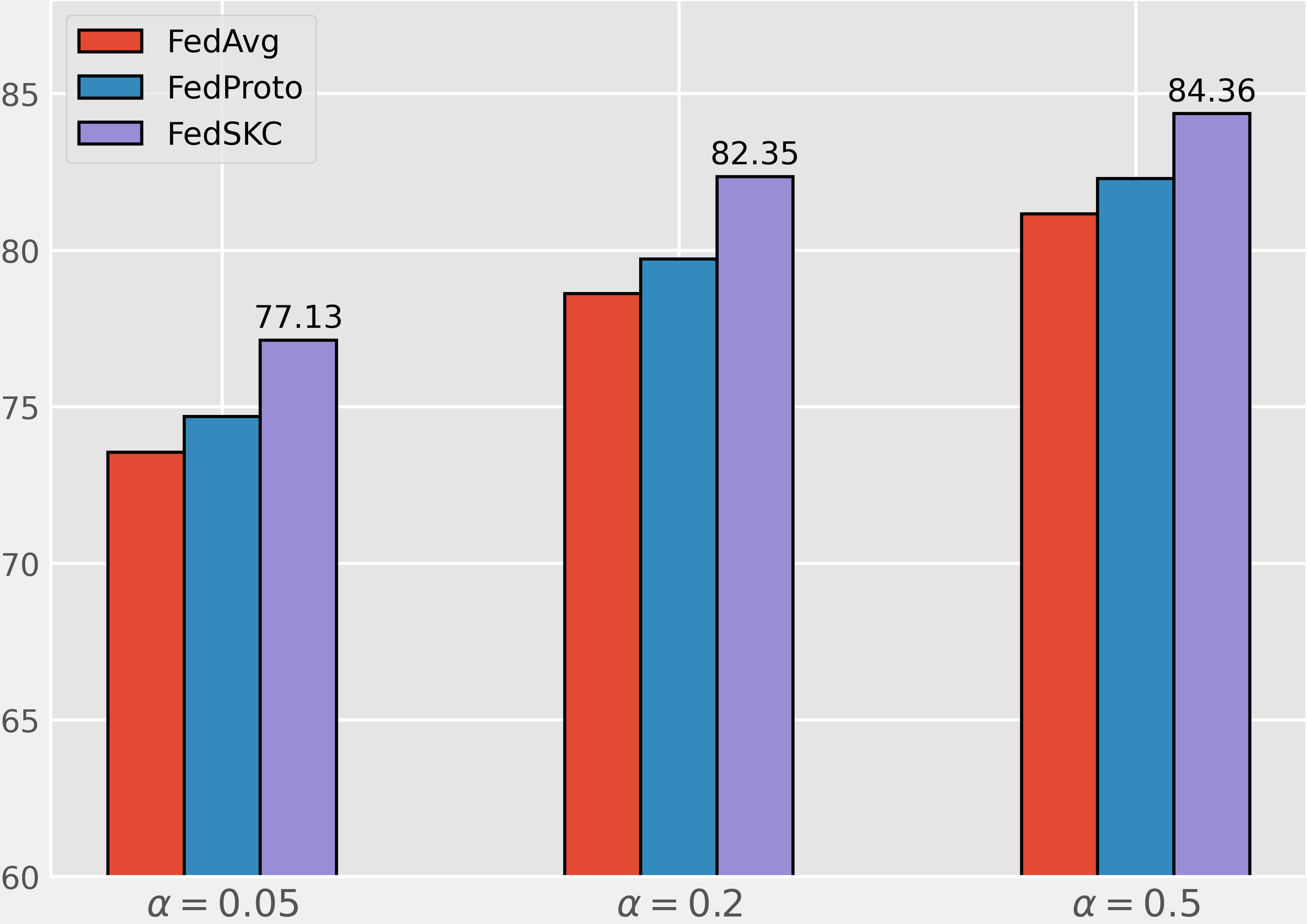}}
    \end{minipage}
    \begin{minipage}[t]{0.245\linewidth}
        \centering
        \subfigure[Imbalanced Level $\rho$]{\label{fig:fig3d}\includegraphics[width=0.98\linewidth]{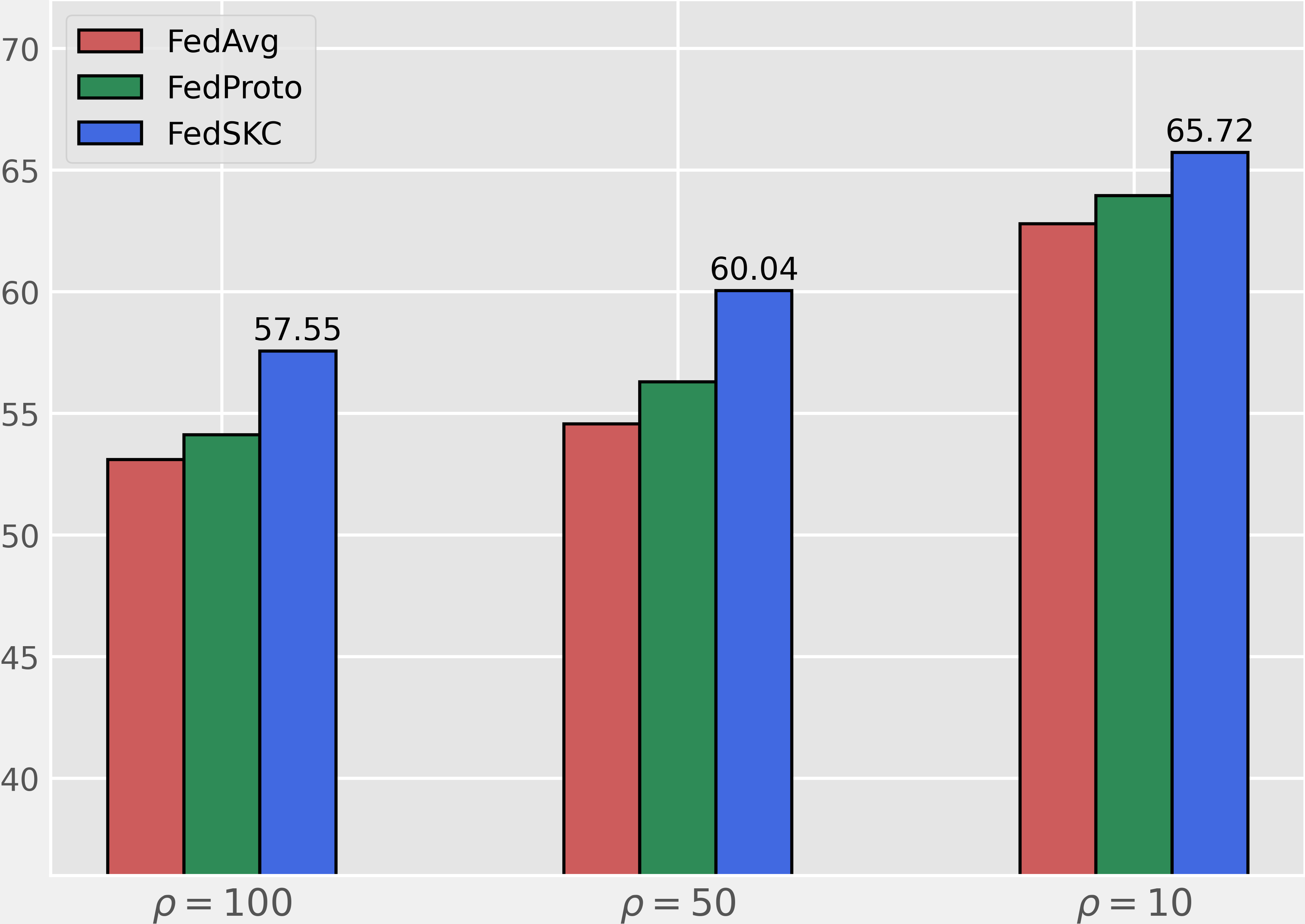}}
    \end{minipage}
    \caption{Effects of four key arguments ($\tau$ in Eq.~\eqref{eq3}, $\mathcal{M}$ in Eq.~\eqref{eq1}, $\alpha$, and $\rho$) in our proposed FedSKC.}
    \label{fig:fig3}
\end{figure*}

\begin{figure*}[tbp]
    \centering
    \begin{minipage}[t]{0.262\linewidth}
        \centering
        \subfigure[t-SNE by FedSKC]{\label{fig:fig4a}\includegraphics[width=1\linewidth]{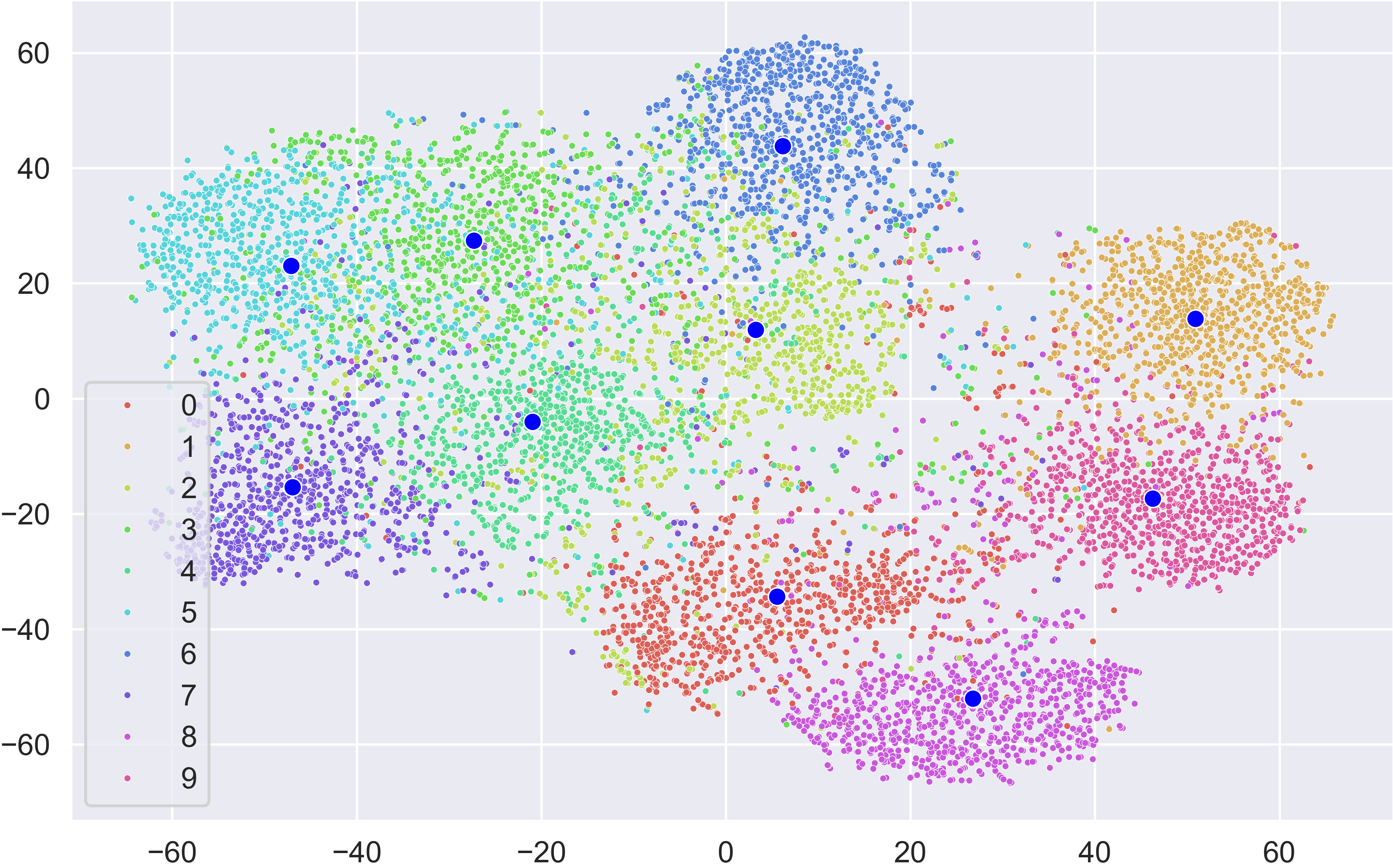}}
    \end{minipage}
    \begin{minipage}[t]{0.26\linewidth}
        \centering
        \subfigure[t-SNE by FedAvg]{\label{fig:fig4b}\includegraphics[width=1\linewidth]{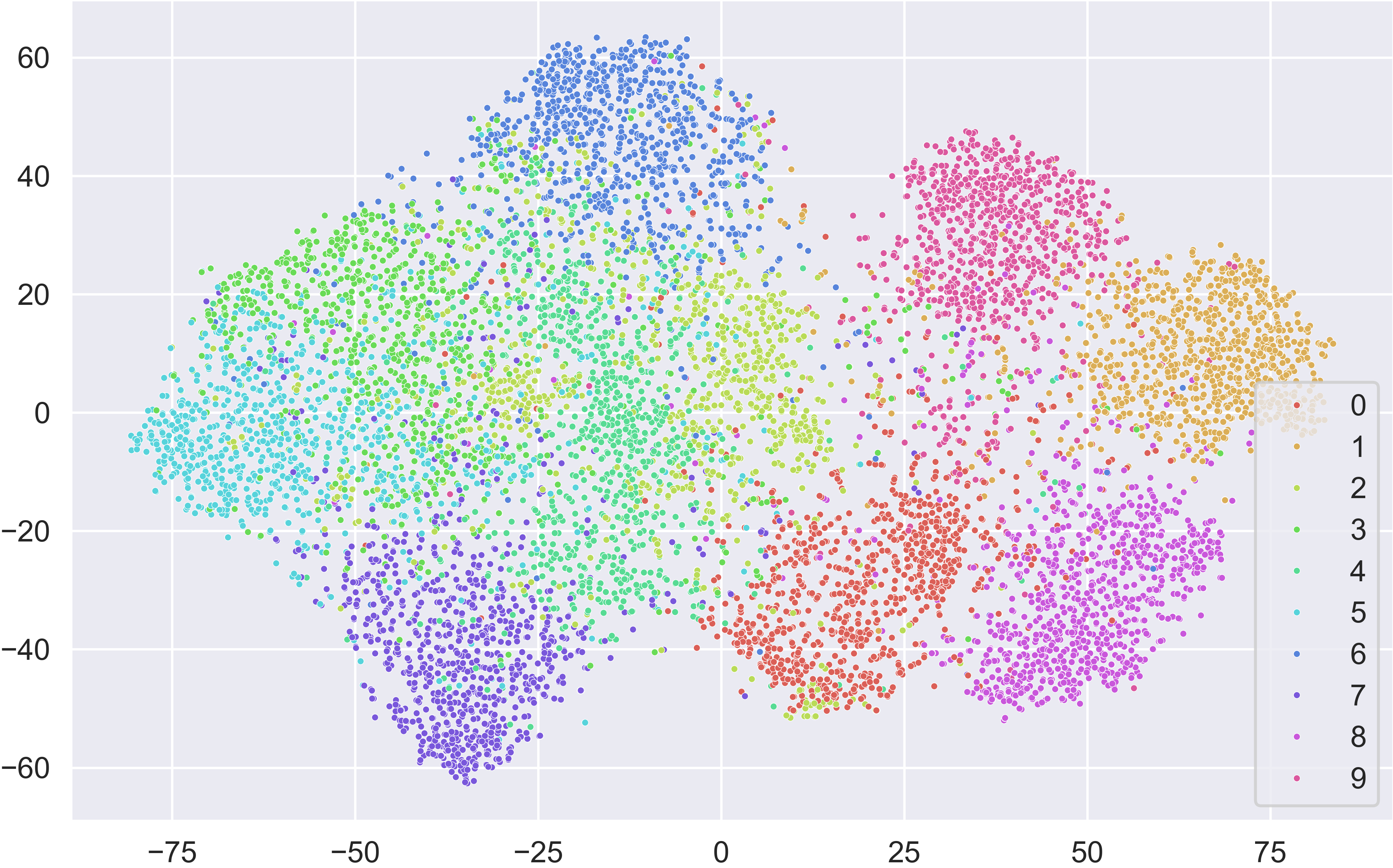}}
    \end{minipage}
    \begin{minipage}[t]{0.23\linewidth}
        \centering
        \subfigure[Effect of Eq.\eqref{eq1}]{\label{fig:fig4c}\includegraphics[width=1\textwidth]{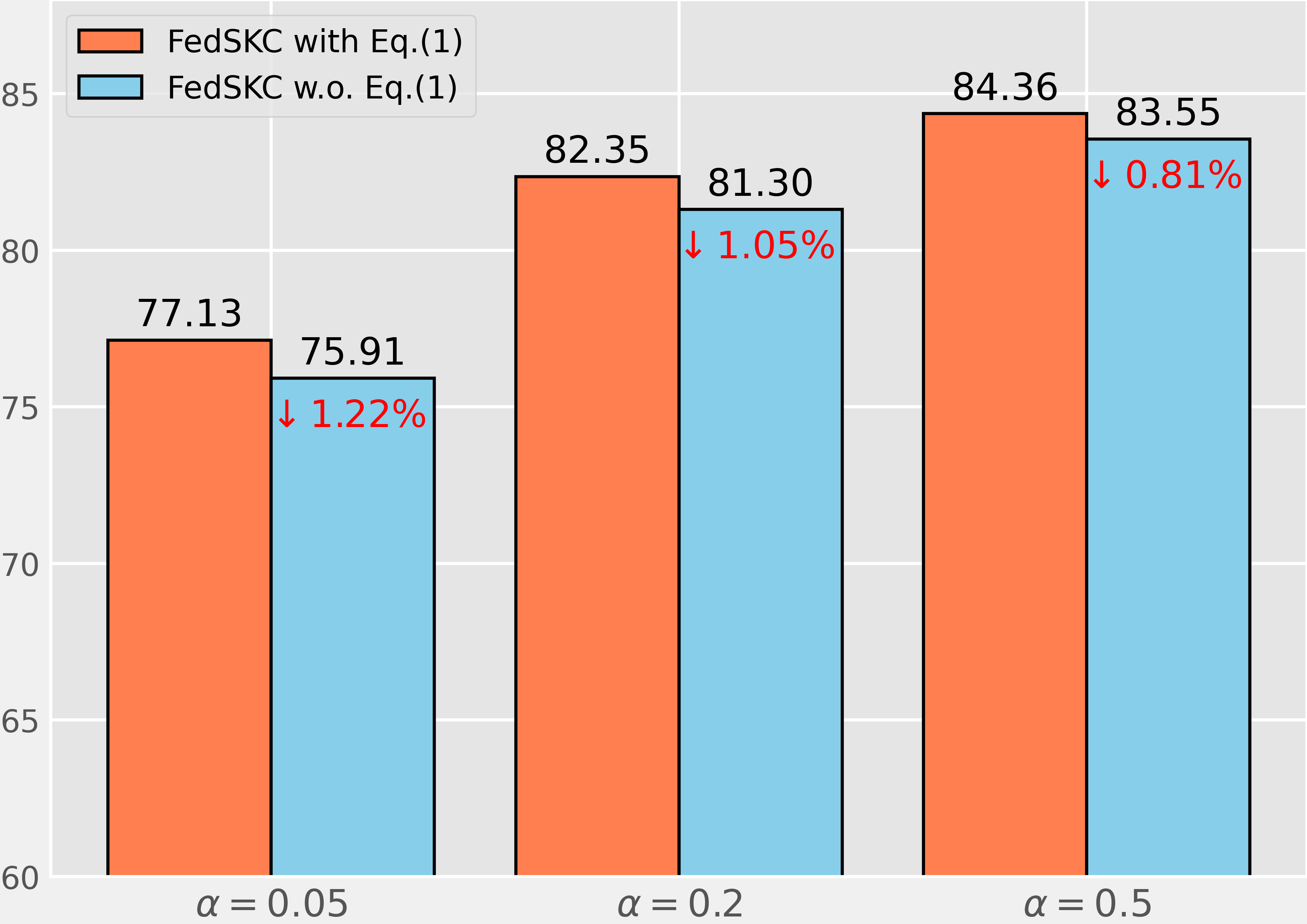}}
    \end{minipage}
    \begin{minipage}[t]{0.22\linewidth}
        \centering
        \subfigure[Illustration]{\label{fig:fig4d}\includegraphics[width=0.95\linewidth]{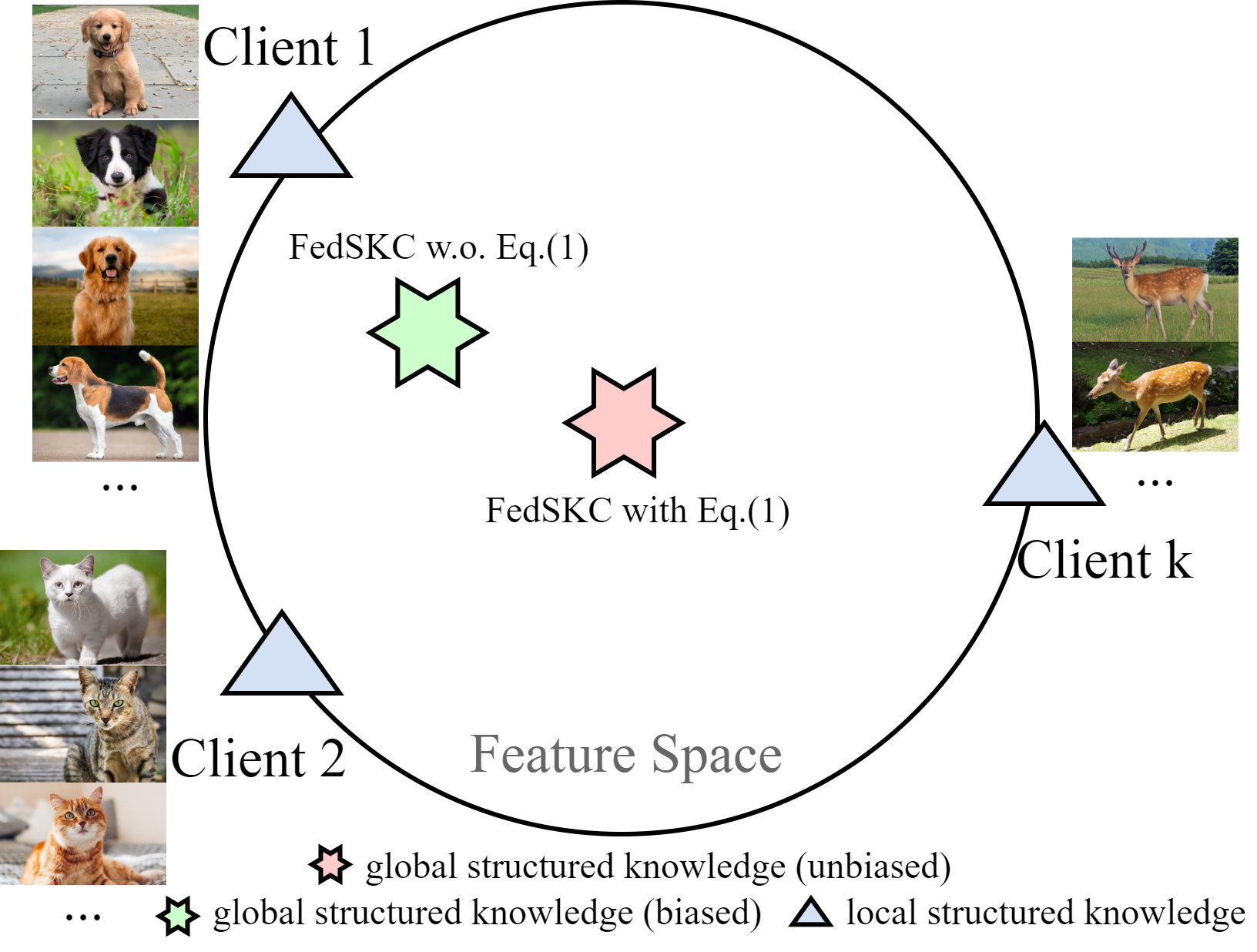}}
    \end{minipage}
    \caption{Analysis for the local and global structural knowledge (Sec.~\ref{sec31}) in our proposed FedSKC.}
    \label{fig:fig4}
\end{figure*}

\subsection{Validation Analysis}
To thoroughly analyze the efficacy of essential modules in FedSKC, we perform validation analysis as follow points:

\textbf{Influence of the temperature $\tau$ in Eq.~\eqref{eq3}:} Following the experiments in Fig.~\ref{fig:fig3a}, the results with different temperature $\tau$ in Eq.~\eqref{eq3}, we can observe that: \textbf{\romannumeral1)} A smaller temperature $\tau$ benefits local training more than higher ones, but an extremely low temperature cannot make model converge ($\mathcal{L}_{LCL}=NaN$ when $\tau=0.01$) due to the training instability and excessive penalties of negative samples, which is confirmed by insights in relevant literature~\cite{wu2024understanding,wang2021understanding,zhang2022dual}; \textbf{\romannumeral2)} $\mathcal{L}_{LCL}$ is not sensitive to the temperature $\tau$ when $\tau \in [0.07,0.11]$, showing the effectiveness of normalization factor $\mathcal{U}$ in Eq.~\eqref{eq2}, which focuses on relative distances among different sample pairs to match the similarity. FedSKC reaches the optimal test accuracy when $\tau=0.08$.

\textbf{Influence of the similar clients $\mathcal{M}$ in Eq.~\eqref{eq1}:} We validate the effects of the number of similar clients ($\mathcal{M}$ in Eq.~\eqref{eq1}) on global structural knowledge, as illustrated in Fig.~\ref{fig:fig3b}. Results show that: \textbf{\romannumeral1)} To calculate global structural knowledge, merging local structural knowledge from the nearest client via Eq.~\eqref{eq1} is better than the single one ($\mathcal{M}=0$), which also reflects that Eq.~\eqref{eq1} helps to enhance the expression of the global structural knowledge; \textbf{\romannumeral2)} An excessively high $\mathcal{M}$ with FedSKC leads to poor performance, showing that information from other clients is not universally beneficial, we choose $\mathcal{M}=1$ by default.

\textbf{Effects of heterogeneity level $\alpha$ and imbalanced level $\rho$:} We tune two key parameters, including the heterogeneity level ($\alpha \in \{0.05, 0.2, 0.5\}$, smaller $\alpha$ denotes more heterogeneity), and the imbalanced level ($\rho \in \{100, 50, 10\}$, larger $\rho$ denotes more imbalanced), results are shown in Fig.~\ref{fig:fig3c} and Fig.~\ref{fig:fig3d}. Experiments indicate that FedSKC consistently exhibits better performance under different FL scenarios. Moreover, the larger $\rho$ introduces more accuracy degradation of the baselines than smaller $\alpha$, showing that most FL methods only deal with data heterogeneity without taking global label imbalance, FedSKC improves this via the GDA module with Eq.~\eqref{eq4}.

\textbf{Analysis of the local structural knowledge:} To prove the superiority of local structural knowledge, we visualize test set of CIFAR-10 by t-SNE~\cite{tsne2008}. As shown in Fig.~\ref{fig:fig4a}, Fig.~\ref{fig:fig4b}, FedSKC performs t-SNE visualization based on local model, small points in different colours represent samples in different labels, with large blue points representing corresponding local structural knowledge. We can observe that samples within the same label are close but separable in FedSKC, and the local structural knowledge (large blue point) is typical for semantics.

\textbf{Analysis of the global structural knowledge:} The global structural knowledge can be considered as an effective supervision signal: \textbf{\romannumeral1)} FedSKC merges the nearest client via Eq.~\eqref{eq1} to regulate local training of each client, promoting representation and generalizability; \textbf{\romannumeral2)} FedSKC provides a consistent regularization via unified global structural knowledge ($\tilde{c}$ in Eq.~\eqref{eq3}) for each local model. As shown in Fig.~\ref{fig:fig4c}, we compare the performance of FedSKC without (w.o.) or with Eq.~\eqref{eq1}, results show that global structural knowledge promotes a satisfactory generalizable performance in FL. Illustration of Fig.~\ref{fig:fig4d} also intuitively explains this viewpoint. Ablation study of the LCL module in Tab.~\ref{tab:tab3} also demonstrates the effectiveness of $\tilde{c}$.

\begin{table}[t]
    \caption{The ablation study of FedSKC with the $\alpha=0.2$ setting.}
    \label{tab:tab3}
    \centering
    \resizebox{\linewidth}{!}{
        \begin{tabular}{ccc||cc||cc}
            \toprule[1.0pt] 
            \textbf{LCL} & \textbf{GDA} & \textbf{GPR} & \textbf{CIFAR-10} & $\triangle$ & \textbf{CIFAR-100} & $\triangle$ \\
            \midrule\midrule 
                      &          &          & 78.61 & - & 55.67 & - \\
            \ding{51} &          &          & 80.45 & \textbf{+ 1.84} & 56.83 & \textbf{+ 1.16} \\
            \ding{51} &\ding{51} &          & 81.63 & \textbf{+ 3.02} & 57.71 & \textbf{+ 2.04} \\
            \ding{51} &          &\ding{51} & 81.28 & \textbf{+ 2.67} & 57.40 & \textbf{+ 1.73} \\
            \midrule[0.66pt]
            \rowcolor{myblue}
            \ding{51} &\ding{51} &\ding{51} & \textbf{82.35} & \textbf{+ 3.74} & \textbf{58.15} & \textbf{+ 2.48} \\
            \bottomrule[1.0pt]
        \end{tabular}
    }
\end{table}

\textbf{Ablation study of FedSKC:} Tab.~\ref{tab:tab3} exhibits the effectiveness and importance of FedSKC modules, the first row refers to FedAvg without extra operation, $\triangle$ denotes the accuracy gain compared to the FedAvg. Experiments show that: \textbf{\romannumeral1)} LCL module leads to significant performance improvements against FedAvg, proving that the global structural knowledge promotes a clear class-wise decision boundary for local model training via Eq.~\eqref{eq3}; \textbf{\romannumeral2)} We notice accuracy gains by introducing GDA and GPR, confirming the discrepancy between local and global structural knowledge is beneficial to aggregation via Eq.~\eqref{eq4}, and also indicating the importance in regard to fine-tuning the global model via Eq.~\eqref{eq5}; \textbf{\romannumeral3)} Combining LCL, GDA, and GPR achieves optimal performance, which supports our motivation to utilize local and global structural knowledge across clients to regulate model optimization during FL training.

\section{Conclusion}
In this paper, we decompose the data heterogeneity problem into local, global, and sampling drift sub-problems, and propose a novel FL framework FedSKC with structural knowledge collaboration. FedSKC has three key components (LCL, GDA, and GPR) to tackle the challenging FL scenarios by leveraging complementary advantages from the local and global structural knowledge: fruitful class-relevant information and stable convergence signals. The effectiveness of the proposed FedSKC has been comprehensively analyzed from both theoretical and experimental perspectives under different FL scenarios.

\section*{Acknowledgment}
This work is supported by the scholarship from the China Scholarship Council (CSC) while the first author pursues his PhD degree in the University of Wollongong. This work was also partially supported by the Australian Research Council (ARC) Linkage Project under Grant LP210300009. This work was partially supported by the Fundamental Research Funds for the Central Universities under Grant No.ZYTS24138.

\setcounter{table}{0}
\setcounter{figure}{0}
\setcounter{equation}{0}
\setcounter{theorem}{0}
\appendix

\subsection{Theoretical Analysis of FedSKC}\label{app:ana}
Here, we provide insights into the convergence analysis for FedSKC under non-convex objectives. The detailed description notations (Sec.~\ref{app:a1}), assumptions (Sec.~\ref{app:a2}), proofs (Sec.~\ref{app:a3}).

\subsubsection{Preliminaries}\label{app:a1}
We introduce some additional variables to better represent the FL training process and model updates from a mathematical perspective. For the $k$-th client, based on Sec.~\ref{sec31}, define the local model $f_{k}(w_{k}): \mathbb{R}^{d_{x}} \rightarrow \mathbb{R}^{d_{\mathcal{C}}}$, and $\mathcal{D}^{k}=\left\{x_{i}, y_{i}\right\}_{i=1}^{N_{k}}$ is the training dataset for the $k$-th client, $N_{k}$ is the number of total samples, label $y_{i}$ of single sample belongs to one of $\mathcal{C}$ classes and is denoted as $y_{i}\in\{1,\ldots,\mathcal{C}\}$. In our FedSKC process, each client $k$ will calculate the local structural knowledge $c_{k}=\{c_{k}^{1},\ldots,c_{k}^{\mathcal{C}}\}$ after local training is completed, then send $f_{k}(w_{k})$ and $c_{k}$ to the server. Then the server receives the local structural knowledge and the local model parameters of all clients, performs a series of processes, and finally sends the global model $f_{g}^{r}$ and global structural knowledge $\tilde{c}_{r}=\{\tilde{c}^{1}_{r},\ldots,\tilde{c}^{\mathcal{C}}_{r}\}$ to active clients on round $r$.

\subsubsection{Assumptions}\label{app:a2}
For this theoretical analysis of FedSKC is based on the following four standard assumptions in FL:
\begin{assumption}[Smoothness]
    \label{assumption1}
    For the objective function $\mathcal{L}$ is $L_{1}$-Lipschitz smooth, 
    which suggests the gradient of objective function $\mathcal{L}$ is $L_{1}$-Lipschitz continuous: 
    \begin{equation} \label{eq:app1}
    \begin{aligned}
        & \| \nabla\mathcal{L}_{k,r_{1}} - \nabla\mathcal{L}_{k,r_{2}}\|_{2} \leq 
        L_{1} \| w_{k,r_{1}} - w_{k,r_{2}}\|_{2}, \\
        & \forall \; r_{1},r_{2} > 0, \enspace k \in \{1,\ldots,K\},
    \end{aligned}
    \end{equation}
    since the smoothness depends on the derivative of the differentiable function, 
    it also indicates the following quadratic bound (there exists the $L_{1} > 0$):
    \begin{equation} \label{eq:app2}
    \begin{aligned}
        \mathcal{L}_{k,r_{1}} - \mathcal{L}_{k,r_{2}} \leq & 
        (\nabla\mathcal{L}_{k,r_{2}})^{\top} (w_{k,r_{1}} - w_{k,r_{2}}) \\
        & + \frac{L_{1}}{2} \| w_{k,r_{1}} - w_{k,r_{2}}\|_{2}^{2}.
    \end{aligned}
    \end{equation}
\end{assumption}
\begin{assumption}[Unbiased Gradient and Bounded Variance]
    \label{assumption2}
    For each client, stochastic gradient is unbiased ($\zeta$ is the sample of the client dataset), so the unbiased gradient is expressed as:
    \begin{equation} \label{eq:app3}
    \begin{aligned}
        & \mathbb{E}_{\zeta}[\nabla \mathcal{L}(w_{k,r}, \zeta)] = \nabla \mathcal{L}(w_{k,r}) = \nabla \mathcal{L}_{r}, \\
        & \forall \; k \in \{1,\ldots,K\}, \enspace r>0,
    \end{aligned}
    \end{equation}
    and has bounded variance: 
    \begin{equation} \label{eq:app4}
    \begin{aligned}
        & \mathbb{E}_{\zeta}[\| \nabla \mathcal{L}(w_{k,r}, \zeta) - \nabla \mathcal{L}(w_{k,r}) \|_{2}^{2}] \leq 
        \sigma^{2}, \\
        & \forall \; k \in \{1,\ldots,K\}, \enspace \sigma^{2} \geq 0.
    \end{aligned}
    \end{equation}
\end{assumption}
\begin{assumption}[Bounded Dissimilarity of Stochastic Gradients]
    \label{assumption3}
    For each objective function $\mathcal{L}$, 
    there exists constant $B$ such that the stochastic gradient is bounded as:
    \begin{equation} \label{eq:app5}
    \begin{aligned}
        & \mathbb{E}_{\zeta}[\| \nabla \mathcal{L}(w_{k,r}, \zeta) \|_{2}] \leq B, \\
        & \forall \; k \in \{1,\ldots,K\}, \enspace B>0, \enspace r>0.
    \end{aligned}
    \end{equation}
\end{assumption}
\begin{assumption}[Lipschitz Continuity]
    \label{assumption4}
    Each real-valued function $f(w)$ is $L_{2}$-Lipschitz continuous:
    \begin{equation} \label{eq:app6}
    \begin{aligned}
        & \| f_{k,r_{1}}(w_{k,r_{1}}) - f_{k,r_{2}}(w_{k,r_{2}}) \| \leq 
        L_{2} \| w_{k,r_{1}} - w_{k,r_{2}}\|_{2}, \\
        & \forall \; r_{1},r_{2} > 0, \enspace k \in \{1,\ldots,K\}.
    \end{aligned}
    \end{equation}
\end{assumption}

\subsubsection{Proofs}\label{app:a3}
In this subsection, we present the three theorems (Sec.~\ref{sec33}) and prove them as follows:

Theorem~\ref{theorem1} is for an arbitrary client, therefore we omit client notation $k$, and we define gradient descent as $w_{r+1} = w_{r} - \eta \nabla f_{r}(w_{r}) = w_{r} - \eta \theta_{r}$, then we can get:
\begin{equation}
        \begin{aligned}
            \mathcal{L}_{rE+2} 
            & \overset{\text{(\romannumeral1)}}{\leq} \mathcal{L}_{rE+1} + (\nabla\mathcal{L}_{rE+1})^{\top} 
                (w_{rE+2} - w_{rE+1}) \\
                & \quad \quad + \frac{L_{1}}{2} \| w_{rE+2} - w_{rE+1} \|_{2}^{2} \\
            & = \mathcal{L}_{rE+1} - \eta (\nabla\mathcal{L}_{rE+1})^{\top} \theta_{rE+1} + 
                \frac{L_{1}}{2} \| \eta \theta_{rE+1} \|_{2}^{2},
    \end{aligned}
\end{equation}
where (\romannumeral1) from the $L_{1}$-Lipschitz quadratic bound of Assumption~\ref{assumption1}. Then, we perform the expectation on both sides of the above equation function as:
\begin{equation}
        \begin{aligned}
            \mathbb{E}[\mathcal{L}_{rE+2}] 
            & \overset{\text{(\romannumeral1)}}{\leq} \mathcal{L}_{rE+1} - \eta \mathbb{E}[\| \nabla\mathcal{L}_{rE+1} \|_{2}^{2}] + 
                \frac{L_{1}}{2} \eta^{2} \mathbb{E}[\| \theta_{rE+1} \|_{2}^{2}] \\ 
            & \overset{\text{(\romannumeral2)}}{\leq} \mathcal{L}_{rE+1} - \eta \mathbb{E}[\| \nabla\mathcal{L}_{rE+1} \|_{2}^{2}] + 
                \frac{L_{1}}{2} \eta^{2} \\ 
                & \quad \quad (\mathbb{E}^{2}[\| \nabla\mathcal{L}_{rE+1} \|_{2}] + \textnormal{Variance} [\| \theta_{rE+1} \|_{2}] ) \\
            & = \mathcal{L}_{rE+1} - \eta \| \nabla\mathcal{L}_{rE+1} \|_{2}^{2} + \frac{L_{1}}{2} \eta^{2} 
                    (\| \nabla\mathcal{L}_{rE+1} \|_{2}^{2} \\
                & \quad \quad + \textnormal{Variance} [\| \theta_{rE+1} \|_{2}]) \\
            & = \mathcal{L}_{rE+1} + (\frac{L_{1}}{2} \eta^{2} - \eta) \| \nabla\mathcal{L}_{rE+1} \|_{2}^{2} \\
                & \quad \quad + \frac{L_{1}}{2} \eta^{2} \textnormal{Variance} [\| \theta_{rE+1} \|_{2}] \\
            & \overset{\text{(\romannumeral3)}}{\leq} \mathcal{L}_{rE+1} + (\frac{L_{1}}{2} \eta^{2} - \eta) 
                    \|\nabla\mathcal{L}_{rE+1}\|_{2}^{2} + \frac{L_{1}}{2} \eta^{2} \sigma^{2},
    \end{aligned}
\end{equation}
where (\romannumeral1) from the stochastic gradient is unbiased in Eq.~\eqref{eq:app3}; (\romannumeral2) from the formula $\textnormal{Variance}(x)=\mathbb{E}[x^{2}]-\mathbb{E}^{2}[x]$; (\romannumeral3) from the Eq.~\eqref{eq:app4}. Then, we telescope $E$ steps on both sides of the above equation to achieve the $rE+E=(r+1)E$ step, which can be expressed as follows:
\begin{equation} \label{eq:app9}
        \begin{aligned}
            \mathbb{E}[\mathcal{L}_{(r+1)E}] 
            & \leq \mathcal{L}_{rE+1} + (\frac{L_{1}}{2} \eta^{2} - \eta) 
            \sum_{\mathbf{e}=1}^{E-1} \|\nabla\mathcal{L}_{rE+\mathbf{e}}\|_{2}^{2} \\
            & \quad \quad + \frac{L_{1}}{2}E\eta^{2}\sigma^{2},
    \end{aligned}
\end{equation}
then, we consider the relationship between $\mathcal{L}_{rE+1}$ and $\mathcal{L}_{rE}$:
\begin{equation*}
\begin{aligned}
    \mathcal{L}_{(r+1)E+1} 
            & = \mathcal{L}_{(r+1)E} + \mathcal{L}_{(r+1)E+1} - \mathcal{L}_{(r+1)E} \\ 
            & = \mathcal{L}_{(r+1)E} + (\mathcal{L}_{CE,(r+1)E+1}+\mathcal{L}_{LCL,(r+1)E+1}) \\ 
            & \quad - (\mathcal{L}_{CE,(r+1)E}+\mathcal{L}_{LCL,(r+1)E}) \\
            & = \mathcal{L}_{(r+1)E} + \mathcal{L}_{LCL,(r+1)E+1} - \mathcal{L}_{LCL,(r+1)E} \\
            & \overset{\text{(\romannumeral1)}}{\leq} \mathcal{L}_{(r+1)E} + \sum_{j=1}^{\mathcal{C}}(
                    s(\zeta,\tilde{c}_{r+2}^{j}) - s(\zeta,\tilde{c}_{r+1}^{j})
                ) \\
            & \overset{\text{(\romannumeral2)}}{\leq} \mathcal{L}_{(r+1)E} + \sum_{j=1}^{\mathcal{C}}(
                    \| \tilde{c}_{r+2}^{j} \|_{2} - \| \tilde{c}_{r+1}^{j} \|_{2}
                ) \\
            & \overset{\text{(\romannumeral3)}}{\leq} \mathcal{L}_{(r+1)E} + \sum_{j=1}^{\mathcal{C}}(
                    \| \tilde{c}_{r+2}^{j} - \tilde{c}_{r+1}^{j} \|_{2}
                ) \\
            & \leq \mathcal{L}_{(r+1)E} + |\mathcal{C}|(\| \tilde{c}_{r+2} - \tilde{c}_{r+1} \|_{2}) \\
            & \overset{\text{(\romannumeral4)}}{=} \mathcal{L}_{(r+1)E} + |\mathcal{C}|( \| \sum_{k=1}^{K}\frac{N_{k}}{N}\frac{1}{|\mathcal{M}|+1} f(w_{k,(r+1)E};\zeta) \\
                & \quad \quad - \sum_{k=1}^{K}\frac{N_{k}}{N}\frac{1}{|\mathcal{M}|+1} f(w_{k,rE};\zeta) \|_{2}) 
\end{aligned}
\end{equation*}
\begin{equation}
        \begin{aligned} 
            & = \mathcal{L}_{(r+1)E} + |\mathcal{C}|\sum_{k=1}^{K}\frac{N_{k}}{N}\frac{1}{|\mathcal{M}|+1}(
                    \| f(w_{k,(r+1)E};\zeta) \\ 
                & \quad \quad - f(w_{k,rE};\zeta) \|_{2}) \\
            & \overset{\text{(\romannumeral5)}}{\leq} \mathcal{L}_{(r+1)E}+
                |\mathcal{C}| L_{2} \sum_{k=1}^{K}\frac{N_{k}}{N}\frac{1}{|\mathcal{M}|+1}( \\
                & \quad \quad \| w_{k,(r+1)E} - w_{k,rE} \|_{2}) \\
            & = \mathcal{L}_{(r+1)E}+ \frac{|\mathcal{C}| L_{2} \eta}{|\mathcal{M}|+1} \sum_{k=1}^{K}\frac{N_{k}}{N} 
                \| \sum_{\mathbf{e}=1}^{E-1} \theta_{k,rE+\mathbf{e}} \|_{2} \\
            & \overset{\text{(\romannumeral6)}}{\leq} \mathcal{L}_{(r+1)E}+ \frac{|\mathcal{C}| L_{2} \eta}
                    {|\mathcal{M}|+1} \sum_{k=1}^{K}\frac{N_{k}}{N} 
                    \sum_{\mathbf{e}=1}^{E-1} \| \theta_{k,rE+\mathbf{e}} \|_{2},
    \end{aligned}
\end{equation}
where (\romannumeral1) from the $\mathcal{L}_{LCL}$ objective function; (\romannumeral2) from the loss function in Sec.~\ref{sec32} Eq.~\eqref{eq2}: $s(x,\tilde{c})=(\frac{1}{\mathcal{U}} \cdot \frac{x}{\|x\|_{2}} \cdot \frac{1}{\|\tilde{c}\|_{2}})\tilde{c} \leq \tilde{c}$, we take expectation to get $\mathbb{E}[s(x,\tilde{c})] \leq \|\tilde{c}\|_{2}$; (\romannumeral3) from the formula as $\|x\|_{2} - \|y\|_{2} \leq \|x-y\|_{2}$; (\romannumeral4) details from Sec.~\ref{sec31} to calculate global structural knowledge, $\zeta$ is the sample of the client dataset; (\romannumeral5) from the Eq.~\eqref{eq:app6}; (\romannumeral6) from the formula as $\| \sum{x} \|_{2} \leq \sum{\| x \|_{2}}$. We apply the expectation on both sides:
\begin{equation} \label{eq:app11}
        \begin{aligned}
            \mathbb{E}[\mathcal{L}_{(r+1)E+1}] 
            & \leq \mathcal{L}_{(r+1)E} + \frac{|\mathcal{C}| L_{2} \eta}
            {|\mathcal{M}|+1} \sum_{\mathbf{e}=1}^{E-1} \mathbb{E}[\| \theta_{k,rE+\mathbf{e}} \|_{2}] \\
            & \overset{\text{(\romannumeral1)}}{\leq} \mathcal{L}_{(r+1)E} + \frac{|\mathcal{C}| L_{2} \eta EB}{|\mathcal{M}|+1},
    \end{aligned}
\end{equation}
where (\romannumeral1) from the Eq.~\eqref{eq:app5} of Assumption~\ref{assumption3}. Through the above Equation~\eqref{eq:app11}, we can get:
\begin{equation} \label{eq:app12}
        \mathbb{E}[\mathcal{L}_{rE+1}] \leq 
        \mathcal{L}_{rE} + \frac{|\mathcal{C}| L_{2} \eta EB}{|\mathcal{M}|+1},
\end{equation}
and then, based on Eq.~\eqref{eq:app9} and Eq.~\eqref{eq:app12}, take the expectation on both sides, we can get:
\begin{equation} \label{eq:app13}
        \begin{aligned}
            \mathbb{E}[\mathcal{L}_{(r+1)E}] 
            & \leq 
            \mathcal{L}_{rE} + \frac{|\mathcal{C}| L_{2} \eta EB}{|\mathcal{M}|+1} 
            + (\frac{L_{1}}{2} \eta^{2} - \eta) \\
            & \quad \quad \sum_{\mathbf{e}=1}^{E-1} \|\nabla\mathcal{L}_{rE+\mathbf{e}}\|_{2}^{2} + \frac{L_{1}}{2}E\eta^{2}\sigma^{2} \\
            & \overset{\text{(\romannumeral1)}}{\leq} \mathcal{L}_{rE} - 
                (\eta - \frac{L_{1}}{2} \eta^{2})EB^{2} + \\
            & \quad \quad \frac{L_{1}E\eta^{2}}{2}\sigma^{2} + 
                \frac{L_{2} E \eta |\mathcal{C}| B}{|\mathcal{M}|+1},
    \end{aligned}
\end{equation}
where (\romannumeral1) from the Eq.~\eqref{eq:app5} of Assumption~\ref{assumption3} and $\mathbb{E}[\|\nabla\mathcal{L}_{r}\|_{2}] \equiv \mathbb{E}[\|\nabla f_{r}(w_{r})\|_{2}]$. \textit{We have completed the proof of \textbf{Theorem~\ref{theorem1}}}.

Based on the Theorem~\ref{theorem1} from Eq.~\eqref{eq:app13}, we can easily get:
\begin{equation}
        \mathbb{E}[\mathcal{L}_{(r+1)E}] - \mathcal{L}_{rE} \leq (\frac{L_{1}}{2} \eta^{2} - \eta) EB^{2} + 
                \frac{L_{1}E\eta^{2}\sigma^{2}}{2} + 
                \frac{L_{2} E \eta |\mathcal{C}| B}{|\mathcal{M}|+1}.
\end{equation}
Then, to ensure the right side of the above equation $- (\eta - \frac{L_{1}}{2} \eta^{2}) EB^{2} + \frac{L_{1}E\eta^{2}}{2}\sigma^{2} + \frac{L_{2} E \eta |\mathcal{C}| B}{|\mathcal{M}|+1} \leq 0$, we can easily get the following condition for $\eta$:
\begin{equation}
        \begin{aligned}
        \eta_{\mathbf{e}} < \frac{2(|\mathcal{M}|+1)B^{2}-2L_{2}|\mathcal{C}|B}{L_{1}(|\mathcal{M}|+1)(\sigma^{2}+B^{2})} 
        \quad \textnormal{where} \enspace \mathbf{e}=1,2,\ldots,E-1.
    \end{aligned}
\end{equation}
So, the convergence of $\mathcal{L}$ holds as round $r$ increases and the limitation of $\eta$. \textit{We have completed the proof of \textbf{Theorem~\ref{theorem2}}}.

Based on the Eq.~\eqref{eq:app13}, we can get:
\begin{equation}
        \begin{aligned}
            \mathbb{E}[\mathcal{L}_{(r+1)E}] 
            & \leq 
            \mathcal{L}_{rE} + \frac{|\mathcal{C}| L_{2} \eta EB}{|\mathcal{M}|+1} 
            + \frac{L_{1}}{2}E\eta^{2}\sigma^{2} \\
            & \quad \quad + (\frac{L_{1}}{2} \eta^{2} - \eta) \sum_{\mathbf{e}=1}^{E-1} \|\nabla\mathcal{L}_{rE+\mathbf{e}}\|_{2}^{2}, 
    \end{aligned}
\end{equation}
and the following equation:
\begin{equation}
        \begin{aligned}
            \sum_{\mathbf{e}=1}^{E-1} \|\nabla\mathcal{L}_{rE+\mathbf{e}}\|_{2}^{2} 
            & \leq \frac{\frac{2|\mathcal{C}| L_{2} \eta EB}{|\mathcal{M}|+1} +
                        L_{1}E\eta^{2}\sigma^{2}}{2\eta - L_{1}\eta^{2}} \\
            & \quad + \frac{2(\mathcal{L}_{rE}-\mathbb{E}[\mathcal{L}_{(r+1)E}])}{2\eta - L_{1}\eta^{2}}.
    \end{aligned}
\end{equation}
Let the round $r$ from $0$ to $R-1$ and step $\mathbf{e}$ from $1$ to $E$:
\begin{equation}
        \begin{aligned}
            \frac{1}{RE} \sum_{r=0}^{R-1} \sum_{\mathbf{e}=1}^{E-1} 
            \mathbb{E}[\|\nabla\mathcal{L}_{rE+\mathbf{e}}\|_{2}^{2}] 
            & \leq \frac{2\sum_{r=0}^{R-1}(\mathcal{L}_{rE}-\mathbb{E}[\mathcal{L}_{(r+1)E}])}{RE(2\eta - L_{1}\eta^{2})} \\
            &  \quad + \frac{\frac{2|\mathcal{C}| L_{2} \eta REB}{|\mathcal{M}|+1} + 
            L_{1}RE\eta^{2}\sigma^{2}}{RE(2\eta - L_{1}\eta^{2})},
    \end{aligned}
\end{equation}
given any $\xi>0$, $\mathcal{L}$ and $R$ receive the following limitation:
\begin{equation} 
    \frac{\frac{2}{RE}\sum_{r=0}^{R-1}(\mathcal{L}_{rE}-\mathbb{E}[\mathcal{L}_{(r+1)E}]) + 
    \frac{2|\mathcal{C}| L_{2} \eta B}{|\mathcal{M}|+1} +
    L_{1}\eta^{2}\sigma^{2}
    }{2\eta - L_{1}\eta^{2}} < {\xi}.
\end{equation}
Then, we can easily get the following condition for $R$:
\begin{equation} 
    R > 
    \frac{2(|\mathcal{M}|+1)(\mathcal{L}_{0}-\mathcal{L}^{*})}
    {\Omega_{1} - \Omega_{2} - \Omega_{3}},
\end{equation}
where $\mathcal{L}^{*}$ denotes the optimal solution of $\mathcal{L}$, $\Omega_{1}=\xi E \eta (|\mathcal{M}|+1)(2-L_{1}\eta)$, $\Omega_{2}=(|\mathcal{M}|+1)L_{1}E\eta^{2}\sigma^{2}$, $\Omega_{3}=2L_{2}E\eta|\mathcal{C}|B$, and $\sum_{r=0}^{R-1}(\mathcal{L}_{rE}-\mathbb{E}[\mathcal{L}_{(r+1)E}])= \mathcal{L}_{0} - \mathcal{L}_{1} + \mathcal{L}_{1} - \mathcal{L}_{2} + \cdots + \mathcal{L}_{R-1} - \mathcal{L}_{R} = \mathcal{L}_{0} - \mathcal{L}_{R} \leq \mathcal{L}_{0} - \mathcal{L}^{*}$. Then, the denominator of the above equation should be greater than 0, so we can easily get the following condition for $\eta$:
\begin{equation} 
        {\eta} < \frac{2 \xi (|\mathcal{M}|+1) - 2 L_{2}|\mathcal{C}|B}
                    {L_{1}(|\mathcal{M}|+1)(\xi + \sigma^{2})}.
\end{equation}
\textit{We have completed the proof of \textbf{Theorem~\ref{theorem3}}}.

\textbf{Summary:} The Theorem~\ref{theorem1} indicates the deviation bound of the objective function $\mathcal{L}$ for an arbitrary client after each round. Convergence can be guaranteed by choosing an appropriate $\eta$. The Theorem~\ref{theorem2} is to ensure the expected deviation of $\mathcal{L}$ to be negative, so the FedSKC's objective function converges, which can guide the choice of appropriate values for the learning rate $\eta$ to guarantee the convergence. Finally, Theorem~\ref{theorem3} provides the convergence rate for FedSKC, the smaller $\xi$ is, the larger $R$ is, which means that the tighter the bound is, the more communication rounds $R$ is required.

\bibliographystyle{IEEEtran}
\bibliography{main}
\end{document}